\pdfoutput=1

\documentclass[11pt]{article}

\usepackage[final]{emnlp2023}

\usepackage{times}
\usepackage{latexsym}

\usepackage[T1]{fontenc}

\usepackage[utf8]{inputenc}

\usepackage{microtype}
\usepackage{graphicx}
\usepackage{url}
\usepackage{booktabs}
\usepackage{multirow}
\usepackage{multicol}
\usepackage{geometry}
\usepackage{tikz}
\usetikzlibrary{shapes,backgrounds}
\usepackage{xcolor}
\usepackage{amsmath,amssymb}
\usepackage{subcaption}
\usepackage{float}

\newcommand{\dataset}{GapChat}

\title{Mind the Gap Between Conversations \\ for Improved Long-Term Dialogue Generation}
\author{
        Qiang Zhang, \textbf{Jason Naradowsky}, \textbf{Yusuke Miyao} \\
        Department of Computer Science \\ The University of Tokyo \\       \{\href{mailto:qiangzhang714@is.s.u-tokyo.ac.jp}{qiangzhang714}, \href{mailto:narad@is.s.u-tokyo.ac.jp}{narad}, \href{mailto:yusuke@is.s.u-tokyo.ac.jp}{yusuke}\}@is.s.u-tokyo.ac.jp
        }

\begin{document}
\maketitle

\begin{abstract}
Knowing how to end and resume conversations over time is a natural part of communication, allowing for discussions to span weeks, months, or years.  The duration of gaps between conversations dictates which topics are relevant and which questions to ask, and dialogue systems which do not explicitly model time may generate responses that are unnatural.  In this work we explore the idea of making dialogue models aware of time, and present \dataset{}, a multi-session dialogue dataset in which the time between each session varies.  While the dataset is constructed in real-time, progress on events in speakers' lives is simulated in order to create realistic dialogues occurring across a long timespan. We expose time information to the model and compare different representations of time and event progress.  In human evaluation we show that time-aware models perform better in metrics that judge the relevance of the chosen topics and the information gained from the conversation.
\end{abstract}


\section{Introduction}
\begin{figure}[ht!]
    \centering
    \includegraphics[width=0.48\textwidth]{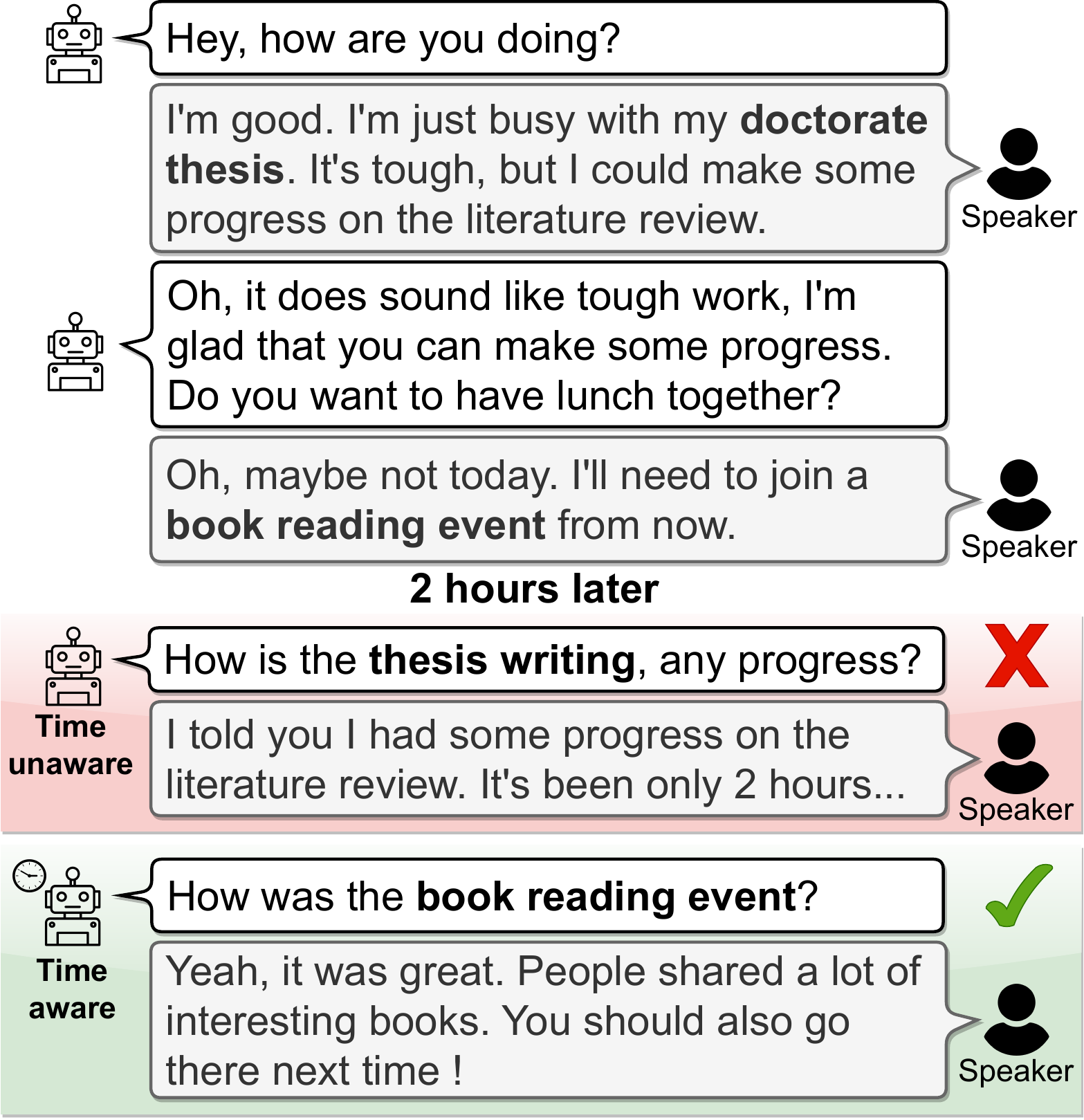}
    \caption{An illustration of the time aware dialogue models we propose. The time aware model talks about events that are informative and effective in a conversation considering the duration of the event and the gap between sessions.}
    \label{fig:intro}
\end{figure}

As language models scale to unprecedented sizes, so too has their ability to generate reasonable and fluent responses to human dialogue.  However, one limitation of large language models (LLMs) towards creating human-like dialogue lies in the fundamental distinction that human cognition is \emph{embodied}, while LLMs are not.  Any considerations which stem from the human physical experience may not be well-represented explicitly in the training data dialogues of LLMs, as reiterating information known to both parties would violate the Gricean maxim of quantity~\cite{grice1975logic}, and may not be effectively utilized during generation.

In this paper we focus on one aspect of embodied experience -- time -- and the role that awareness of time plays in shaping realistic dialogue in multi-session conversations.  
We argue that an awareness of the passage of time between conversations allows human speakers to more accurately gauge which topics and questions will result in new information, leading to conversations which are more informative and appear more natural.  For example (Figure~\ref{fig:intro}), a dissertation is a time-consuming endeavor, so if given a two-hour gap between conversations, it is unlikely that significant progress has been made.  In the context of potential \emph{information gain}, asking for a progress update may be less useful than asking about an event with more expected progress within that time frame, such as a lunch or a meeting.  To enable dialogue models to behave similarly, we propose incorporating an awareness of time as it pertains to (1) how much time has passed since the previous conversation, (2) what events were previously discussed, and (3) what progress of each event is expected over that duration.

Previous work introduced the concept of multi-session chat (MSC) and dataset for the task where the discussion between two speakers is divided with gaps between sessions~\cite{xu-etal-2022-beyond}.  However, gaps between sessions are relatively short (1-7 hours, or 1-7 days), and event topics are derived from basic persona attributes (\emph{I have six cats}) which prevents any discussion of long-term events or event progress. Moreover, annotators vary between sessions, which may have further reduced consistency. 

To remedy these issues and support long-term MSC research, we present ~\dataset{}, a dataset which extends the MSC dataset with additional sessions~\footnote{\url{https://github.com/QZx7/MindTheTime/tree/main}}.  
However, while chat in MSC is open-ended, conversations in~\dataset{} are based on simulated timelines.  This design choice allows for two participants to create a realistic and consistent conversation about long-term events, with gaps between sessions on the scale of days, weeks, or months, in a comparatively shorter period of time.

We explore multiple ways of making dialogue models time-aware, using information such as the ongoing events of speakers, the expected duration of the events, and the duration of time gap between sessions. Our contributions are as follows: (1) the creation of a new MSC-style dataset that simulates explicit gaps in communication, across various time-scales, to enable research into long-term MSC dialogue generation, (2) the exploration of different time-aware dialogue models, which incorporate time information into their contexts to enable varying degrees of time-based reasoning about discourse topics, and (3) we demonstrate via human evaluations of generated dialogues that the inclusion of time information improves the naturalness, informativeness, and relevance of conversations.

\section{Related Work}

\paragraph{Pragmatics Theory of Communication}
Underlying this work is the cooperative principle of communication~\cite{grice1975logic}, in which communication is goal-oriented, and where effective conversations are those which adhere to basic principles, including the desire to communicate only information which requires it. The cooperative principle has been used as a means to evaluate both human-human~\cite{eskritt2008preschoolers,kleinke2010speaker} and human-computer~\cite{lan2020pone,langevin2021heuristic} conversations. To assess the quality of follow-up questions posed by artificial agents,~\citeauthor{ge2022should} (\citeyear{ge2022should}) proposed measures based on Gricean maxims to capture aspects such as relevance, informativeness, truthfulness, clarity, and coherence. Motivated by this line of work, we explore the role that time awareness plays in gaining information effectively by avoiding discussing non-mentionable topics.

\paragraph{Long-term Dialogue Model}
Research on long-term dialogue generation based on neural networks has primarily focused on means of storing and accessing long-term information beyond the bounds of a limited context window.  One approach to this problem is selecting valuable information for long-term storage, such as conversation summaries~\cite{xu-etal-2022-beyond} and persona attributes ~\cite{xu-etal-2022-long}. This approach involves training a model which can retrieve from storage information which is relevant to the current topic. An alternative to retrieval, attention mechanisms can fulfill a similar role~\cite{zhang-etal-2022-history}. 
\begin{figure*}[ht!]
    \centering
    \includegraphics[width=\textwidth]{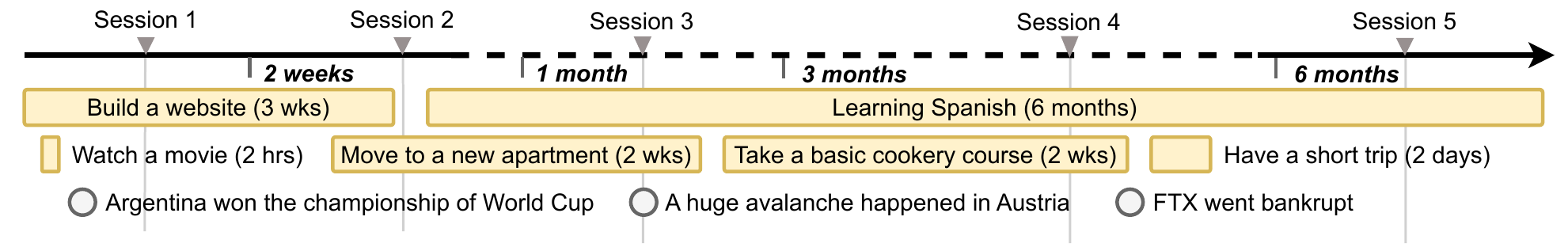}
    \caption{An example of the simulated timelines for a speaker in \dataset{}. Each speaker will engage in specific events that take a certain period of time to finish. Yellow rectangles and circles indicate life events and grey circles indicate world events.}
    \label{fig:timeline}
\end{figure*}

\paragraph{Dialogue Models on Topic Selection}
A potential benefit of time-awareness is topic selection that results in more informative conversations.  Previous approaches have used semantic relevance as the basis for topic selection~\cite{somasundaran2020two,xu2021topic}, or have selected based on topic transition patterns~\cite{xie2021tiage,ling2021context}.
While such factors are shown to be effective in general discourse, in the context of multi-session discourse with gaps between conversations, commonsense about previous topics may help rule out topics that are unlikely to result in new information being discussed.  If this is a guiding principle in human discourse, conversations that utilize temporal information should appear more natural.

\section{\dataset{}: A Time-Aware MSC Dataset}

The MSC dataset contains gaps between sessions and their durations are annotated.  However, the gaps were short, and the annotators were changed between sessions. They were not instructed to pay attention to the relation between topics, events and the duration between gaps. T
hus temporal information in MSC has been shown to have a minimal effect on model performance~\cite{xu-etal-2022-beyond}. 
However, we hypothesize that the use of a dataset with discussions focusing on more realistic long-term events may have a more significant impact on model performance.  
To test this hypothesis, we create a new conversation dataset, \dataset{}.  
We follow the general settings in MSC and similarly collect dialogue data via crowdsourcing.  Thus \dataset{} can also be used as additional data for any task utilizing MSC.  
However, unlike MSC, conversations in \dataset{} are not completely open-ended, but are grounded to the hypothetical lives of the two annotators, who are each given a procedurally generated timeline of events in their lives.  This notion of a simulated life provides speakers with updates to events in their lives which are realistic with respect to the time which is said to have passed since the last chat session.  
Sessions are then scheduled to take place randomly at various positions along the timeline, which may occur before, during, or after any particular event.

\subsection{Events and Timelines}
\label{sec:timelines}
We define a \emph{timeline} as a linear sequence of (possibly overlapping) events, where each event is represented by a string label (\emph{watch a movie}) together with an expected duration (\emph{2 hours}).  When two annotators are selected, a series of events is sampled, and timelines created for both speakers by randomly ordering the sampled events.  There are two types of events: life events and world events.

\paragraph{Life Events}
A life event is an event that is said to happen in the person's life and they are exclusive to each speaker.  We use life events to ensure the continuity in the multi-session conversations. Life events are crafted in two ways: manually, and generated with ChatGPT~\footnote{\url{https://platform.openai.com/docs/models/gpt-3-5}}. When crafting events manually, we collect example life events from online resources~\footnote{\url{https://simplicable.com/philosophy/life-events}; \url{https://www.dudleycourtpress.com/50-life-events-for-your-own-memoir/}}. The durations of these events are estimated by searching online forums (e.g., Quora, Reddit) using the query ``How long will it usually take to finish <event>?''. Up to 5 answers are selected, and the average estimation is used as the duration.

To include a more diverse range of events in our dataset, we also use ChatGPT to generate additional life events that vary in duration and domain. 
When generating with ChatGPT, we prompt the model to ``Generate a list of events or daily activities that require around <duration> to finish,'' where <duration> represents the time needed to complete the event (e.g., 1-3 weeks). 
Manually and generated, a total of 50 life events with varying durations, including hours, days, weeks, months, and years are collected, with 10 events for each duration. 

In order to reflect changes in the topic over time, longer events are further subdivided into a series of steps denoted as an event schedule.  To do that, we ask ChatGPT to generate the steps towards finishing each event via prompting (see Figure~\ref{fig:craft_events}). For instance, in the case of attending a 3-day basic cookery course, the introduction may require 1 day to complete, learning knife skills may take another day, and learning to cook basic dishes may take 2 more days. And these three steps make up a schedule for the event of ``attending a basic cookery course''. A dialogue session could randomly be assigned to occur at any place within this event, or not at all. We create event schedules for each event that is with a duration that is longer than ``1 hour''.  Each schedule of an event consists of a maximum of 7 steps. For events longer than a month, we generate two schedules with varying sets of steps, each representing a different approach to completing these long-term events. Examples of life events and their corresponding schedules can be found in Appendix~\ref{apx_a}.

\paragraph{World Events}
Not all discussion topics pertain to the lives of the speakers, and it is common to also have discussions on current events.  
To account for this and to add further realism to the conversations, we also include world events.  We define a world event as a newsworthy event which takes place in the world, and is thus contained in the timelines of both speakers, such as ``\emph{Argentina wins the World Cup championship}''.   We collected 78 world events by extracting news article titles from Google News between November 2022 and January 2023.  When used in timelines, world events happen in the same order as in the real world.

\begin{table}[t!]
    \centering
    \begin{tabular}{ccc}
    \toprule
    \textbf{\#Session} & \textbf{Dialogues} & \textbf{Utterances}\\
    \midrule
    3 & 150 & \phantom{0}8,235\\
    4 & 300 & 26,651\\
    5 & 200 & 21,368\\
    \midrule
    Total & 650 & 56,254\\ 
    \bottomrule
    \end{tabular}
    \caption{Data statistics of \dataset{}.}
    \label{tab:statistics}
\end{table}

\subsection{Data Collection}
We collect data via Amazon Mechanical Turk (AMT), beginning with the easier task of 3-session conversations and inviting reliable workers to take part in the collection of 4 and 5 session conversations.  A total of 48 workers were selected to participate in the final data collection. We require fluency in English for all participants. Workers are paid \$7 per hour on average.  The instructions and interface used for data collection can be found in Appendix~\ref{apx_c}.

\paragraph{Initial session}
At the beginning of the initial session, each speaker is provided the first event in their timelines (e.g., ``You just started attending a basic cookery course, which would take about 4 days'') as an initial event to share in the conversation. The initial events are randomly selected from the life events. 
Once the minimum session length is reached, either speaker can end the current session, move time forward, and create a new session with a randomly generated time gap ranging from 10 minutes to 1 year. 

\paragraph{Subsequent sessions} 
For all subsequent sessions, speakers are provided with updates on the progress of their ongoing events based on the duration, steps of the events, and the time gap. If the time gap is shorter than the minimum required time to reach the next step in the schedule of the event, speakers will receive a message stating ``No significant progress.''  
Conversely, if the time gap covers the full duration of the event, speakers will be notified that the event has been completed and new life events will be provided according to the timeline. 

\begin{figure}[t]
    \centering
    \includegraphics[width=0.48\textwidth]{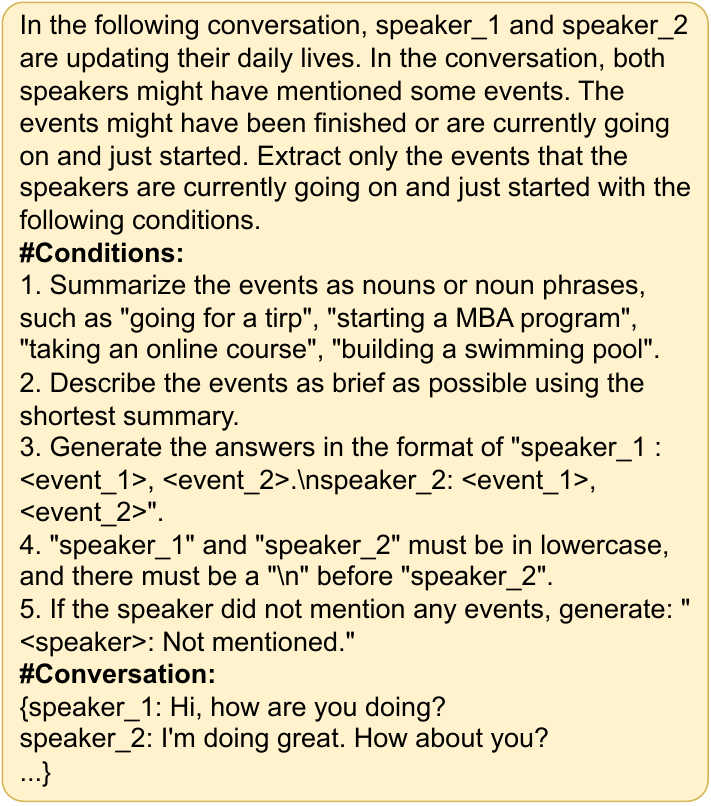}
    \caption{The prompt we use to extract events from a given conversation during inference.}
    \label{fig:extract_event}
\end{figure}
\paragraph{Session details \& statistics}
A typical MSC session contains 12-14 utterances, and this may make it difficult for speakers to engage in in-depth discussions before transitioning to a new session.  We observe that many sessions in MSC immediately continue the conversation of the previous session, essentially making the data no different than a longer single session.  
To address this, we design each session in \dataset{} to have at least 20 utterances and to start and end in a natural way (e.g., greetings and closures), and we encourage sessions to continue past this minimum. 
Additionally, we instruct speakers to be mindful of the time gap between sessions as it may influence their behavior in the subsequent session.  Table~\ref{tab:statistics} shows the statistics of \dataset{}. In total we collect 56,254 utterances across 2,650 sessions.  Each conversation consists of 3$\sim$5 sessions. 
Examples of a 4-session conversation is available in Appendix~\ref{apx_b}.  For evaluation we split the data to Train/Valid/Test with a ratio of 0.7/0.1/0.2.



\begin{figure*}
    \centering
    \includegraphics[width=\textwidth]{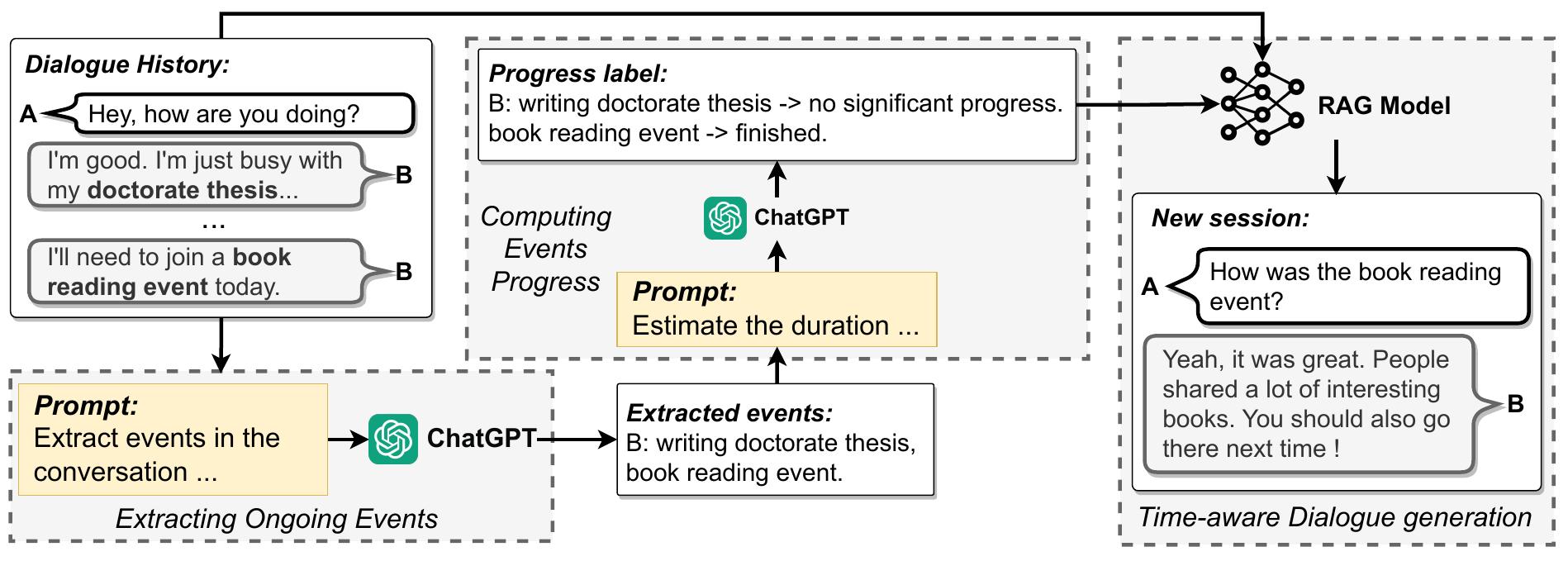}
    \caption{The illustration of the process of modeling time with progress label. When using schedules, an additional step of prompting LLM for schedule information is added.}
    \label{fig:modeling_time}
\end{figure*}

\section{Modeling Time}
\label{sec:model_time}

We propose a method of modeling time in multi-session dialogue systems. 
As shown in Figure~\ref{fig:modeling_time}, this consists of three procedures: 1) extracting ongoing events, 2) computing events progress, and 3) time-aware dialogue generation.

\subsection{Extracting Ongoing Events}
At training time, dialogue events are observed, since they are pre-defined in timelines that are assigned to the speakers. However, at inference time, it is necessary to extract ongoing events on-the-fly from the dialogue history.  We phrase this as a text generation task where the input is the preceding dialogue history and the output is a list of events discussed in the history. The events to extract are defined as in Section~\ref{sec:timelines}. For instance, if a speaker mentioned ``I'm currently working on a research project until the end of the semester'', the event ``working on a research project'' is extracted as a life event.
We extract events in a prompt-based manner using ChatGPT. 
Different styles of prompts are explored, as shown in Table~\ref{tab:completion}$\sim$~\ref{tab:qa_2} in Appendix~\ref{apx_d}. The prompt which resulted in the highest extraction performance is shown in Figure~\ref{fig:extract_event}, and is used in the following experiments.

\begin{figure}[t]
    \centering
    \includegraphics[width=0.48\textwidth]{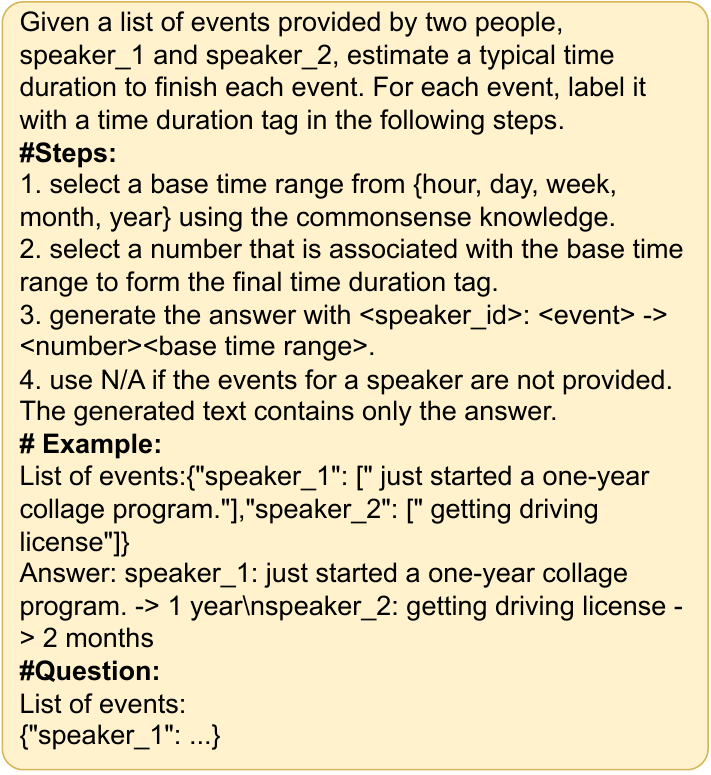}
    \caption{The prompt used to estimate event durations.}
    \label{fig:estimate_duration}
\end{figure}

\subsection{Computing Events Progress}
During training, we compare the event duration against the session gap to determine the progress of events. At inference time, we first estimate the duration of the events extracted in the previous procedure through prompting. We use two styles of methods to represent the progress of the events: the \textit{progress labels} and \textit{event schedules}.

\paragraph{Estimating event duration}
Event duration is estimated by querying a knowledge base that has the temporal commonsense knowledge as in the MCTACO dataset ~\cite{zhou-etal-2019-going}. 
We utilize ChatGPT as a knowledge base, and prompt it to provide a typical duration for each event. Figure~\ref{fig:estimate_duration} shows the prompt we use for estimating event duration. 

\paragraph{Using progress labels}
When using progress labels to represent the event progress, we formulate event progress as a discrete labeling task, where each extracted event in the previous procedure is labelled with one of five progress labels, \{\textit{No significant progress, 1/4 finished, half finished, 3/4 finished, finished}\}. Finer or real-valued estimation of progress is possible, but was not pursued due to potential sparsity issues.  
Event durations are then compared with the time gap to calculate the progress labels for each event. For instance if the duration of the event ``getting a driver license'' is ``2 months'' and the time gap is ``6 weeks'', the event is given a progress label of ``3/4 finished''.
Although it is possible to prompt ChatGPT directly to generate the progress labels, we calculate these labels to prevent potential inaccuracies or hallucinations that ChatGPT might produce regarding mathematical tasks~\cite{bang2023multitask}.

\paragraph{Using event schedules}
As an alternative to the more numerically oriented progress labels, the schedules collected in Section~\ref{sec:timelines} contain steps that are required to finish an event, and thus can also represent the progress of the events. We leverage the schedules and split a schedule into two lists ``finished'' and ``to-do''. The ``finished'' part contains those steps that can be completed during the session gap, and the ``to-do'' part contains the remaining steps. For instance, a schedule of the ``getting the driver license'' contains 5 steps of ``one week for learning rules, 2 weeks for practicing, 2 weeks for passing exams, one week for road check, one week for getting license.'' If the time gap is ``two weeks,'' the ``finished'' list includes ``one week for learning rules,'' and the remaining steps are added to the ``to-do'' list. The ``finished'' part of the schedule represents the current progress of the event.

\subsection{Time-aware Dialogue Generation}
We use a RAG (Retrieval-Augmented Generation) 2.7B model~\cite{lewis2020retrieval} for dialogue generation. A RAG model utilizes a retriever to retrieve related contexts stored as documents in memory, making it effective at handling large collections of text. The truncation of this model is set to 1024, enabling it to encode more context.

\paragraph{Documents and retriever}
Following MSC, we save the dialogue context as documents and retrieve with a DPR model~\cite{karpukhin-etal-2020-dense}. Different sessions are saved as separate documents in the memories encoded by the DPR model. When provided with the dialogue context, the top-5 documents are retrieved for response generation.

\paragraph{Training the model}
During training time, events and different types of time-aware information (progress labels and schedules) are combined with dialogue history as context in a manner similar to Persona-Chat~\cite{zhang-etal-2018-personalizing} (Table~\ref{tab:model_input}).
\begin{table}[t]
    \centering
    \begin{tabular}{l}
    \toprule
    A: Hey, how are you doing? \\
    B: I'm good. I'm just busy with my doctorate\\
    thesis. \\
    ... \\
    B: I'll need to join a book reading event today. \\
    \textbf{Events} \\
    B: writing doctorate thesis, book reading \\
    event. \\
    \textbf{Progress} \\
    B: writing doctorate thesis [no significant\\
    progress], book reading event [finished].\\
    \textbf{Gap} \\
    2 hours \\
    \bottomrule
    \end{tabular}
    \caption{The sample input to time-aware models.}
    \label{tab:model_input}
\end{table}

We train different time-aware RAG models (TA-RAG) by providing various time-aware information (gaps, progress labels and schedules) using the MSC-RAG 2.7B 1024 as the initial model. A baseline model, RAG (FT), is also trained with only the dialogue history and extracted events but without time-aware information. All models are trained on the ParlAI platform~\footnote{\url{https://parl.ai/}} in a seq2seq style, with the dialogue context as the input and the response the label. The models are trained with 2 NVIDIA A100 GPUs for 72 hours. 

\section{Experiments}
Models are separated into two groups, time-unaware models and time-aware models. Human evaluation is conducted to compare all models against RAG (FT).

\begin{table*}[ht!]
    \centering
    \begin{tabular}{crccccc}
    \toprule
    \multicolumn{2}{r}{\textbf{Model}} & \multicolumn{5}{c}{\textbf{Human Ratings against RAG (FT)}} \\
    \midrule
    & & Naturalness & Informativeness & Relevance & Time-Awareness & Total\\
    \cmidrule(lr){3-7}
    \textbf{Time-} & \multicolumn{1}{l}{MSC-RAG}  &   - 3.17  &	  \phantom{0}1.12 &	- 4.68	& - 44.10 &  - 12.71  \\
    \textbf{unaware} & \multicolumn{1}{l}{ChatGPT} & \phantom{0}7.64  &  - 0.50  &   \phantom{0}1.00   &  20.28  &  \phantom{01}7.11 \\
    \midrule
    & \multicolumn{1}{l}{TA-RAG:} \\
    & \textit{progress}      & \phantom{0}9.02  &   \phantom{0}5.62  & 	13.75  &  44.82  &  18.30\\
    & \textit{schedule}  & \phantom{0}9.16  &   \phantom{0}4.06   &	15.22  &  \textbf{54.84}  &  20.82\\
    \textbf{Time-} & \textit{both}      & 15.15   &   \textbf{\phantom{0}6.32}  &   \textbf{ 20.78}  &  50.56  &  \textbf{23.20} \\
    \textbf{aware} & \multicolumn{1}{l}{ChatGPT:} \\
    & \textit{gap only}     & 10.52  &  \phantom{0}2.18  &   \phantom{0}3.12   &  16.66  &  \phantom{0}8.12 \\
    & \textit{progress}      & 18.08  &   \phantom{0}5.12  & 	15.38  &  38.78  &  19.34\\
    & \textit{schedule}  & 15.84  &   \phantom{0}4.54   &	12.30  &   47.78  &  20.12\\
    & \textit{both}      & \textbf{18.26}   &   \phantom{0}3.80  &    17.30  &  52.58  &  22.99 \\
    \bottomrule
    \end{tabular}
    \caption{Human evaluation results of different models when compared against RAG (FT). Negative numbers indicate that the model performs worse than RAG (FT).}
    \label{tab:human_evaluation}
\end{table*}

\subsection{Time-unaware Models}

We include comparisons to a number of models which do not explicitly consider time gap and event duration information as baselines:

\paragraph{MSC-RAG} The RAG 2.7B model proposed by MSC~\cite{xu-etal-2022-beyond}, where the model saves previous dialogue sessions and retrieves relevant information during generation.

\paragraph{RAG (FT)} MSC-RAG model fine-tuned on \dataset{}. No time-aware information is provided in this model.

\subsection{Time-aware Models}

\paragraph{TA-RAG} Time-aware models are based on the MSC-RAG model, fine-tuned on \dataset{}, and using various time-aware information described in Section~\ref{sec:model_time}.  For the TA-RAG (progress) model, the time-aware information is provided in the style of progress labels, and for the TA-RAG (schedule) model, it is in the style of event schedules. TA-RAG (both) indicates that both styles of time-aware information are provided.

\subsection{ChatGPT}
Besides the fine-tuned models, we also add ChatGPT into our experiments to explore its time-awareness. We consider both time-unaware and time-aware scenarios and prompt ChatGPT to generate responses. In the time-unaware case, only dialogue history and extracted events are provided to the model. In the time-aware case, we explore different types of ChatGPT settings. In the ``with gap only'' type, in addition to the dialogue history and events, time gaps between sessions are explicitly given, enabling ChatGPT to recognize the engage in a multi-session conversation. In the ``progress'', ``schedule'' and ``both'' types, progress of events are also provided as in TA-RAG models.


\subsection{Evaluation}

\paragraph{Collecting dialogues}
Models are evaluated using the human evaluation by comparing the conversations they generate. Conversations from all RAG models are generated in a self-chat style, where the RAG models engage in a conversation with a common BlenderBot3B model. Self-chat has been shown to perform comparably to human-model conversations~\cite{smith-etal-2022-human} for evaluation purposes.  In the case of ChatGPT, we utilize prompts to obtain responses (Figure~\ref{fig:chatgpt_prompt}, Appendix~\ref{apx_e}).

Conversations are generated session by session, where the first 3 utterances of the first session are seeded from manually crafted scripts to ensure the models start the conversation by sharing some events.  In subsequent sessions, randomly selected time gaps and new events are provided together with the dialogue history of the previous session to the models. All models use the same time gaps and events to ensure a consistent experiment setting. 

Evaluation follows ACUTE-Eval~\cite{li2019acute}, where annotators are asked to rate the conversations generated by various models in comparison to the baselines. In our experiments, we compare all the models to RAG (FT) and evaluate the conversations session by session. The annotators are provided the dialogue history of previous sessions, the event information and the time gap. They are then asked to select the conversations generated by one model over another. 
For each model, we collect 150 multi-session conversations spanning different session lengths (50 for 3-session, 4-session, and 5-session scenarios, respectively).

\begin{figure*}
    \centering
    \includegraphics[width=0.9\textwidth]{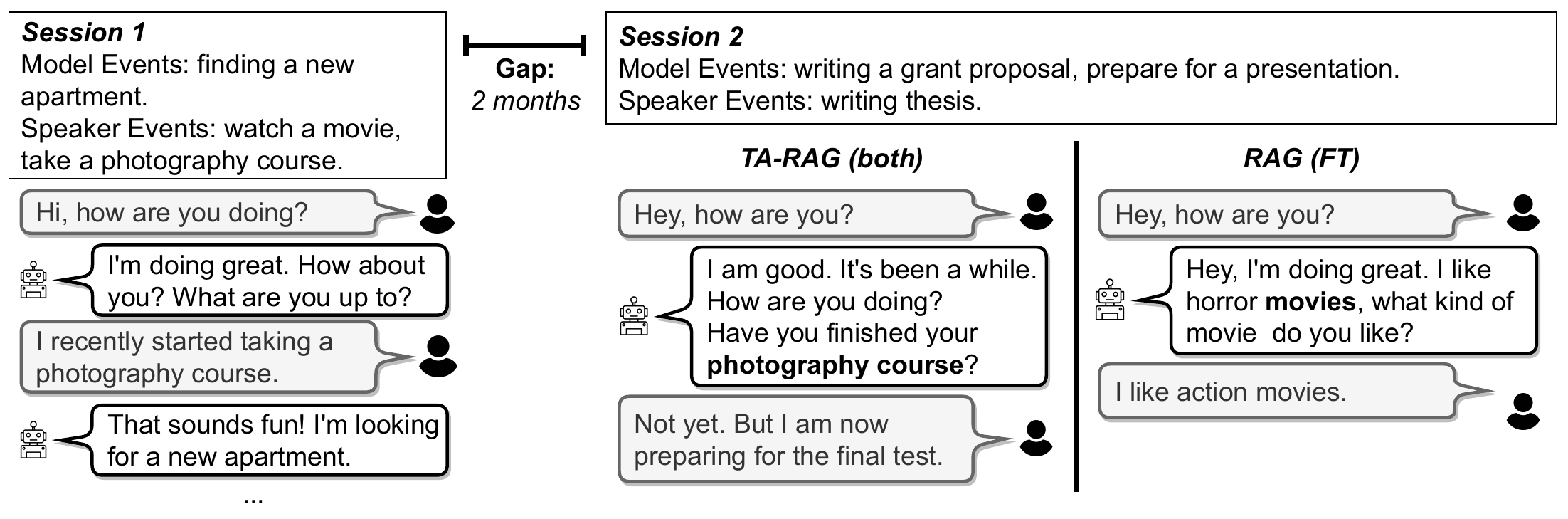}
    \caption{A sample conversation generated by RAG (both) and RAG (FT). The models generate the conversations with the same events and time gap.}
    \label{fig:sample_dialogue}
\end{figure*}

\paragraph{Human evaluation questionnaire}
In the evaluation, annotators are asked a total of 11 questions, grouped as 4 attributes: naturalness, informativeness, relevance and time-awareness. Naturalness refers to the ability of the model to generate conversations that feel like two friends updating each other on their daily lives. Informativeness evaluates whether the model frequently asks frustrating questions about event progress, affecting the information gain~\cite{ge2022should}. Annotators are asked, ``Which speaker asks annoying questions about events with no significant progress?'' Relevance measures the model's ability to generate follow-up sessions related to mentioned events by asking annotators ``In which dialogue does the speaker ask/talk more about relevant events?'' Time-awareness assesses the model's ability to identify time gaps and the progress of events by asking annotators ``In which dialogue does the speaker identify time gaps more accurately?''

We conduct the evaluation on AMT. Questions are phrased differently for each attribute and average ratings are calculated (Appendix~\ref{apx_f}). After pilot tests, 66 annotators are chosen for the final evaluation task. 10 annotators are hired for each task of model comparison. We also request explanations for choices to validate responses. During analysis, we filter out evaluations with short working time (< 200 seconds) and unreasonable explanations (e.g., single-word, repetitive, unrelated to the conversation, or copied from other text). The Fleiss' Kappa across all annotators is $K=73.1\%$.

\section{Results \& Analysis}

\subsection{Main Findings}

Results of the human evaluation are shown in Table~\ref{tab:human_evaluation}.
Our main finding is that \textbf{time information improves overall naturalness of the conversations}, which supports the hypothesis that time-based reasoning of discourse topics is an important part of long-term multi-session dialogues. Figure~\ref{fig:sample_dialogue} shows an example in which TA-RAG (both) is able to produce more natural conversations (more examples in Appendix~\ref{sec:apx_g}).
TA-RAG with one type of time information performs worse than ChatGPT on naturalness, however, overperforms ChatGPT when both types of time information are added.  

Between different types of time information, we find that \textbf{progress labels contribute most to the enhancement of informativeness, but schedule information has relatively minor impact on the informativeness.} This may be attributable to the more direct manner in which this approach provides information to the model, as the progress towards an event represents the outcome of a time-based reasoning process that is performed outside of the model.  Using the progress representation would closely align with a desired discourse action (e.g., do not select events with labels of ``no significant progress''), whereas the schedule representation requires the model to perform additional reasoning.

\subsection{Analysis}

\begin{figure}
    \centering
    \includegraphics[width=0.48\textwidth]{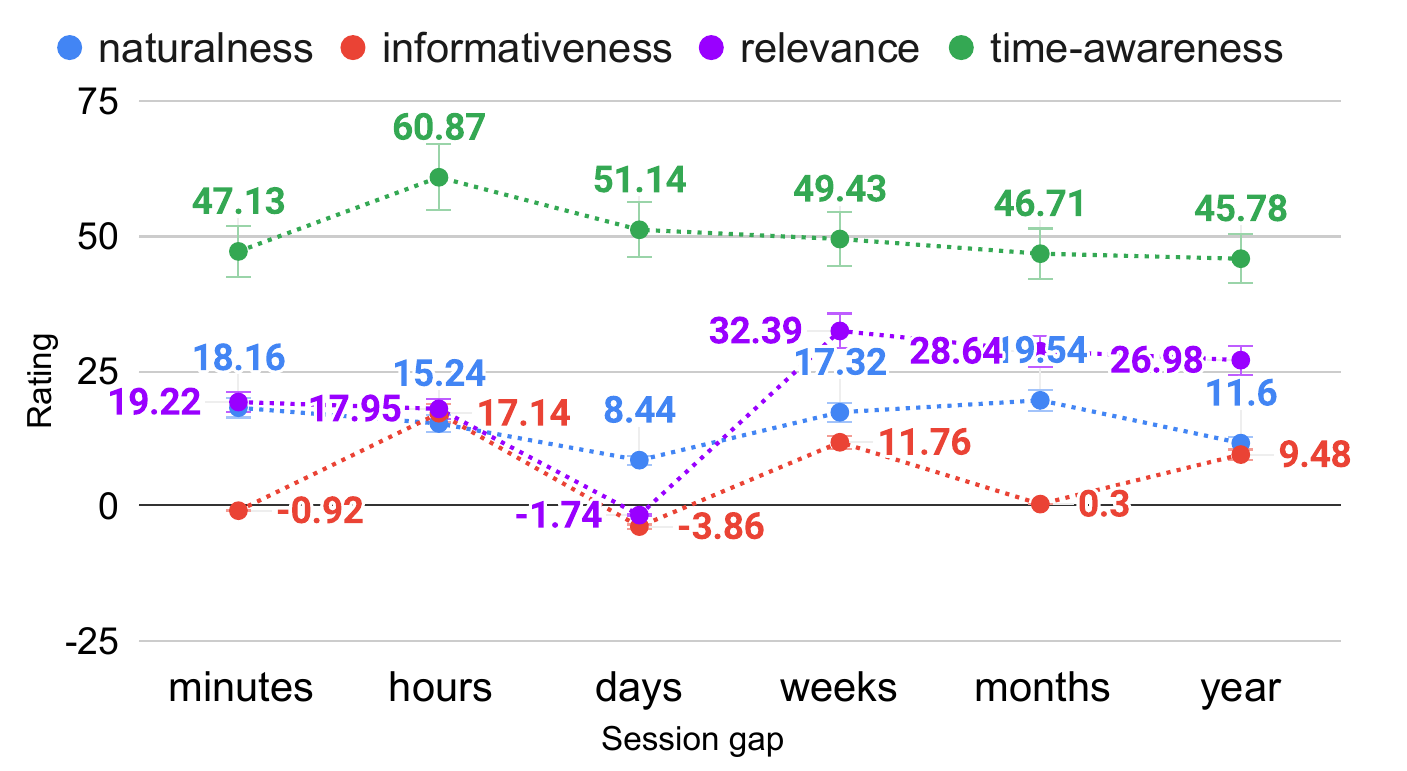}
    \caption{Performance of TA-RAG (both) over different session gaps.}
    \label{fig:session_gap_both}
\end{figure}

\paragraph{Various gap duration}
We analyze the performance of the models with regard to specific gap durations, but do not find a consistent relationship between gap length and the evaluation measures (Appendix~\ref{apx_h}).  In some models (TA-RAG (both)) we find that naturalness and time-awareness are more stable over different session gaps, whereas informativeness and relevance shows larger deviations (see Figure~\ref{fig:session_gap_both}), but these deviations do not change the overall ranking of systems on each attribute.

\paragraph{Time-awareness in ChatGPT}
ChatGPT is considerably larger and trained on more data than any other model in our study, and may have induced a better understanding of time. In fact, we rely on ChatGPT as a source of temporal commonsense when obtaining event durations.  
It is then no surprise that ChatGPT exhibits a certain level of time-awareness, however it does not score as high on this metric as TA-RAG models. Surprisingly, we found that while adding information about the gap between sessions improves naturalness, informativeness and relevance, it does not enhance the time-awareness of ChatGPT. Rather, we observed improved greetings and conclusions, and the inclusion of gaps appears to have primarily improved the overall structure of the dialogue.

\paragraph{ChatGPT outputs lengthy responses resulting in relatively low informativeness.}

To explore the impact of time information on ChatGPT's performance, we use similar prompts, as depicted in Figure~\ref{fig:chatgpt_prompt}, albeit with time information incorporated into the prompts. The inclusion of time information improves time-awareness, demonstrating that this information is useful and not implicitly utilized to the same extent by the original model.  In terms of informativeness, we observe that adding both types of information results in lower performance than only utilizing progress labels.  We find that this is due to the longer length of responses, which annotators naturally consider less informative.  When schedule information is added, ChatGPT exhibits a tendency to incorporate schedule details into the reply, which increases the reply length. 

\paragraph{Adding time information helps the model select events in subsequent sessions.}
We conducted an analysis of the generated conversations to examine the extent to which the TA-RAG models effectively select appropriate events as topics in subsequent sessions. 
\begin{table}[t]
    \centering
    \begin{tabular}{rc}
    \toprule
    \textbf{Model} & \textbf{\#Correct Events} \\
    \midrule
    RAG (FT) & 17 \\
    \midrule
    TA-RAG (progress) &  46  \\
    TA-RAG (schedule) &  42 \\
    TA-RAG (both)     &  44  \\
    \bottomrule
    \end{tabular}
    \caption{Number of session-pairs in which the model selects correct events in a subsequent session. Total number of tested case is 60.}
    \label{tab:following_topics}
\end{table}
We manually review 60 randomly selected consecutive session-pairs (20 for each of 3-session, 4-session and 5-session conversations) and measured the quality of topic selection as the extent the subsequent session avoided addressing events that have a label of ``no significant progress''. 
Table~\ref{tab:following_topics} shows the results of the conversations generated by different models. 
We observe that adding time information, especially the progress label information, results in selection of more appropriate events in subsequent sessions (where the more desirable events are ones which are most likely to cause discussion of new information).

\section{Conclusion}



This work is a study of the role that temporal reasoning plays in shaping dialogues over multiple sessions, and the extent to which temporal information is useful when integrated into existing dialogue systems.  We demonstrate via human evaluations that time-aware models generate text with improved naturalness, informativeness, and relevance, resulting better topic selection and improved text generation quality overall.  The results emphasize the importance for future dialogue models to consider ``time-awareness'' as an important factor in achieving natural conversations. Additionally, our analysis shows that incorporating time-aware information also enhances the performance of existing LLMs such as ChatGPT.

\section*{Limitations}

We rely on existing LLMs (ChatGPT) to serve as knowledgebases for temporal commonsense.  The extracted event durations agreed with author intuitions, but as they are data-driven they may not reflect reality and may give a false expectation of the time necessary to complete events.  The events used in our dataset represent a small set of possible events, and we do not know what coverage this system would have on events in real conversations.  Further, using this system requires ChatGPT in the loop to extract events during inference time, and while accuracy was high and the extracted text was sensible on the events used in this study, there are inherent risks present when using an LLM for text generation, and incorporating that information downstream without human oversight.



\section*{Acknowledgements}
This work was partially supported by JSPS KAKENHI Grant Number 19H05692 and Moonshot R\&D Program Grant Number JPMJMS2237\_4.


\bibliography{anthology,custom}
\bibliographystyle{acl_natbib}

\appendix
\clearpage
\section{Samples of events in simulated timelines}
\label{apx_a}

\begin{table*}[ht]
    \centering
    \begin{tabular}{l}
    \toprule
    \textit{Life event} \\
    \midrule
    You just started preparing and executing a social media marketing campaign for \\
    your company, which would take about 3 months. \\
    \midrule
    \textit{Duration}: 3 months \\
    \midrule
    \textit{Schedule} \\
    \midrule
    \textbf{Steps:} \\
    4 weeks for researching the audience and markets, one week for creating  engaging content that \\
    aligns with the campaign goals, one week for designing and setting up the campaign, 4 weeks for \\
    executing and optimizing the campaign, one week for analyzing the campaign data, one week for \\
    making adjustments to the content and strategies as necessary. \\
    \bottomrule
    \end{tabular}
    \caption{An example of the schedules in simulated timelines.}
    \label{tab:schedule_1}
\end{table*}
\begin{table*}[ht]
    \centering
    \begin{tabular}{l}
    \toprule
    \textit{Life event} \\
    \midrule
    You just started writing your doctor thesis, which would take about one year. \\
    \midrule
    \textit{Duration}: 1 year \\
    \midrule
    \textit{Schedule 1} \\
    \midrule
    \textbf{Steps:} \\
    one month for reviewing the guidelines and outlining the structure of the thesis, 2 months for \\
    writing the introduction, 2 months for writing the remaining chapters, 2 months for revising \\
    the thesis based on feedback from other colleagues, one month for preparing the formatting and \\
    citations, one month for final tough and submitting to committee, one month for addressing \\ 
    comments requested by the committee members and defending the thesis. \\
    \midrule
    \textit{Schedule 2} \\
    \midrule
    \textbf{Steps:} \\
    one month for reviewing the guidelines and outlining the structure of the thesis, 2 months for \\
    writing the introduction, 2 months for writing the remaining chapters, 2 months for revising \\
    the whole thesis for clarity, coherence and flow, one month for revising the thesis based on \\
    feedback, 2 months for preparing and submitting a conference paper, one month for preparing \\
    the extension procedures of the doctorate program. \\
    \bottomrule
    \end{tabular}
    \caption{An example of multiple schedules that lead to different results of the same life event.}
    \label{tab:schedule_2}
\end{table*}

\begin{table*}[t]
    \centering
    \begin{tabular}{l}
    \toprule
    \textit{World events} \\
    \midrule
    Argentina won the championship of World Cup. \\
    Prince Harry reveals whether he’s circumcised in bombshell ‘Spare’ memoir \\
    US closes in on Bankman-Fried Inner Circle with probe of FTX chief engineer. \\
    \bottomrule
    \end{tabular}
    \caption{Examples of world events.}
    \label{tab:world_events}
\end{table*}

Table~\ref{tab:schedule_1} and~\ref{tab:schedule_2} are examples of the life events.
An event can have different schedules that lead to different consequences/ends. We use these schedules to increase the diversity of the directions of developing the ongoing events. For each life event, we pre-define the duration and use ChatGPT to generate steps of a schedule for the event within the duration via prompts. An example prompt can be found in Figure~\ref{fig:get_schedule}. Table~\ref{tab:world_events} shows an example of the world event.


\section{Samples of \dataset{}}
\label{apx_b}

Table~\ref{tab:sample_data_1}$\sim$\ref{tab:sample_data_4} show a 4-session sample conversation we collect. For each session in the dataset, we provide the progress of events and the utterances. For subsequent sessions, we also provide the session gap since the previous session.
\begin{table*}[ht]
    \centering
    \begin{tabular}{l}
    \toprule
    \textit{Session 1} \\
    Events: \\
    Speaker 1: You just started organizing the storage room, which would take about 2 days. \\
    Speaker 2: You are about to have a short vacation to the beach, which would take about 3 days. \\
    \midrule
    Speaker 1: Hi! How are you doing? I heard you're going on a short vacation to the beach. \\ 
    That sounds exciting! \\
    Speaker 2: Hey! Yes, I'm really looking forward to it. It's been a while since I've had a break. \\
    Speaker 1: Definitely, it's always nice to take some time off. Do you have any plans for what\\
    you'll do while you're there? \\
    Speaker 2: Not really, just planning to relax and soak up some sun. Maybe take a dip in the ocean. \\
    Speaker 1: That sounds like the perfect way to unwind. Have you packed everything you need for\\
    the trip? \\
    Speaker 2: Yes, I have everything sorted. I just need to double-check that I haven't forgotten \\
    anything important. \\
    Speaker 1: That's great to hear. On my end, I've been organizing my storage room. It's been a bit \\
    of a mess, so I've been taking some time to sort through everything. \\
    Speaker 2: Oh wow, that sounds like quite the task. How's it going so far? \\
    Speaker 1: It's definitely been time-consuming, but it's starting to look much better. I've been  \\
    able to clear out a lot of clutter and create more space. \\
    Speaker 2: That's great! It always feels good to have a more organized living space. \\
    Speaker 1: Absolutely. Speaking of which, have you been keeping up with your own home \\ 
    organization? \\
    Speaker 2: Yeah, I've been trying to keep things tidy. It's a never-ending process though! \\
    Speaker 1: Tell me about it. It always feels like there's something to clean or organize. Any other \\
    plans for when you return from your vacation? \\
    Speaker 2: Not really, just getting back into the routine of things. How about you? \\
    Speaker 1: I'm planning to work on some writing projects when I finish organizing the storage room. \\ 
    I've been meaning to get back into it for a while. \\
    Speaker 2: Oh, that's interesting! What kind of writing do you do? \\
    Speaker 1: Mostly fiction, but I've been wanting to try my hand at some non-fiction pieces as well. \\
    Speaker 2: That sounds like a lot of fun. Let me know if you need any help or feedback. \\
    Speaker 1: Thanks, I appreciate it! How about we check back in after your vacation and see how \\ 
    everything's been going? \\
    Speaker 2: Sounds good to me! Have a good one. \\
    Speaker 1: You too, enjoy the beach! Talk to you soon. \\
    Speaker 2: Thank you.. \\
    \midrule
    \end{tabular}
    \caption{Session 1 of an sample conversation we collect.}
    \label{tab:sample_data_1}
\end{table*}

\begin{table*}[ht]
    \centering
    \begin{tabular}{l}
    \toprule
    \textit{Session 2} \\
    Gap: 1 week. \\
    Events: \\
    Speaker 1: You just started to prepare for a marathon context, which would take about one year. \\
    Speaker 2: You are about to visit the nearby town for about two days. \\
    \midrule
    Speaker 1: Hey! It's been a week since we last talked. How have you been? \\
    Speaker 2: Oh, hi! I've been great, thanks. How about you? \\
    Speaker 1: I'm doing well, thanks for asking. By the way, how was your beach\\
    vacation? \\
    Speaker 2: It was amazing! The beach was so beautiful and the weather was \\ 
    perfect. I even got a nice tan. \\
    Speaker 1: That's awesome to hear! So, I finished organizing the storage \\ 
    room, and now I've started preparing for a marathon contest that will take \\
    about a year. What about you? Did you have any progress in your recent events? \\
    Speaker 2: That sounds interesting! Well, I've been studying for a certification \\ exam, and I also planned and executed a fundraising campaign for my local \\
    community center. \\
    Speaker 1: Wow, you've been busy! How did the fundraising campaign go? \\
    Speaker 2: It went really well! We managed to raise a good amount of money, \\
    and the community center was very grateful for our help. \\
    Speaker 1: That's great to hear! \\
    Speaker 2: Same great to hear! \\
    Speaker 1: By the way, have you heard about ChatGPT's AI making puzzles \\
    that will make you want to throw brickbats? \\
    Speaker 2: No, I haven't. \\
    Speaker 1: Well, apparently they're really challenging and frustrating, \\
    but also addictive. \\
    Speaker 2: What are these puzzles about? \\
    Speaker 1: Some people are saying they can't stop playing them. \\
    Speaker 2: Oh, I see. That sounds like fun, but also kind of frustrating. \\
    Anyway, what are your future plans for the next 6 months? \\
    Speaker 1: I'm planning to work on a personal project and pursue a passion \\
    that I've been neglecting for a while. How about you? \\
    Speaker 2: I'm planning to spend Monday and Tuesday working on a presentation \\
    that I have to give on Wednesday. \\
    Speaker 1: Good luck with that! \\
    Speaker 2: It's going to be a busy couple of days, but I'm looking forward to it. \\
    Speaker 1: Let me know how it goes. \\
    \midrule
    \end{tabular}
    \caption{Session 2 of an sample conversation we collect.}
    \label{tab:sample_data_2}
\end{table*}

\begin{table*}[ht]
    \centering
    \begin{tabular}{l}
    \toprule
    \textit{Session 3} \\
    Gap: 18 hours. \\
    Events: \\
    Speaker 1: You watch TV dramas. \\
    You play video games.\\
    Speaker 2: Tour the town's main attractions. \\
    Have lunch at a local restaurant. \\
    \midrule
    Speaker 1: Hey, how's it going? \\
    Speaker 2: I've been busy with a lot of things \\
    Speaker 1: It's been almost a day since we last talked. What have you been up to?\\
    Speaker 2: Nothing really exciting to talk about. How about you? \\
    Speaker 1: Well, I've been working on my storage room organization project. \\
    Speaker 2: That's great! \\
    Speaker 1: I finished it yesterday, and now I'm looking forward to starting \\
    something new. \\
    Speaker 2: Congratulations on finishing it. So, what's your next project going to be? \\
    Speaker 1: I'm planning to participate in a marathon that's going to happen next \\ 
    year, so I just started preparing for it. \\
    Speaker 2: Wow, that sounds like a big challenge. \\
    Speaker 1: Yes, my goal is to complete it within a specific time frame. \\
    Speaker 2: Do you have any specific goals for the marathon? \\
    Speaker 1: I'm also planning to set smaller goals for myself, like improving my \\
    running time and endurance. \\
    Speaker 2: That's impressive. Good luck with your training. By the way, did you hear \\ about the latest news on Microsoft and Activision? \\
    Speaker 1: No, what happened? \\
    Speaker 2: Apparently, they backed off from their aggressive claim in the FTC case. \\ It's quite interesting to see how these big companies handle legal issues. \\
    Speaker 1: Hmm, that's interesting indeed. I haven't been keeping up with the news \\ 
    much lately, but I did watch some TV dramas and play some video games during my free\\
    time. \\
    Speaker 2: That's cool. I also spent some time assembling furniture that I bought last\\ week. It was quite challenging, but I'm happy with the end result. \\
    Speaker 1: Oh, nice! What kind of furniture did you assemble? \\
    Speaker 2: It was a bookshelf and a cabinet for my study room. It took me a while to \\ figure out the instructions, but I managed to get it done. \\
    Speaker 1: Good for you! By the way, I'm planning to set aside some time to review \\ 
    and reflect on my personal and professional goals for the year and make any necessary \\ adjustments. Do you have any plans for the next few days? \\
    Speaker 2: Yes, I have to work on a proposal that's due on Friday, so I'm going to\\ spend Wednesday and Thursday on it. How about you? \\
    Speaker 1: I'm planning to choose the marathon and make a commitment to participate.\\
    I'm also planning to set realistic goals for myself and develop a plan to achieve them.\\
    Speaker 2: Sounds like you're making progress. Keep up the good work! \\
    \midrule
    \end{tabular}
    \caption{Session 3 of an sample conversation we collect.}
    \label{tab:sample_data_3}
\end{table*}

\begin{table*}[ht]
    \centering
    \begin{tabular}{l}
    \toprule
    \textit{Session 4} \\
    Gap: 3 hours. \\
    Events: \\
    Speaker 1: You do some housecleaning. \\
    You play with your pet.\\
    Speaker 2: You take a bath. \\
    You do a board game with your family. \\
    \midrule
    Speaker 1: Hey, it's good to talk to you again! \\
    Speaker 2: Hey! It's been an interesting day. \\
    Speaker 1: How have you been spending your time since we last talked? \\
    Speaker 2: I've been watching some TV dramas, playing some video games, and assembling \\ some furniture. What about you? \\
    Speaker 1: Sounds like a nice day. I've been working on my presentation, and it's going \\
    pretty well so far. \\
    Speaker 2: Thanks for letting me know. On my end, I'm planning to set aside some time\\
    to review and reflect on my personal and professional goals for the year. Do you have \\
    any tips on how to do that effectively? \\
    Speaker 1: That's a great idea! One tip I have is to make sure you set aside enough time \\
    for reflection, and really focus on what's most important to you. You might also want to \\
    consider setting some specific, measurable goals to help you achieve your overall \\
    objectives. \\
    Speaker 2: That's a good point. I'm also planning to choose a marathon and make a \\ commitment to participate, as well as set some realistic goals for myself and develop a \\ 
    plan to achieve those goals. Do you have any experience with marathon training? \\
    Speaker 1: Not personally, but I have friends who have trained for marathons before. \\
    It can be a challenging but rewarding experience. \\
    Speaker 2: I'm excited to take on the challenge! By the way, I also wrote a short story \\ this week. Would you like to hear about it? \\
    Speaker 1: Of course! I'm always interested in hearing about your creative projects. \\
    Speaker 2: It's a suspenseful story about a woman who becomes trapped in an elevator with \\ a stranger who may or may not be dangerous.\\
    Speaker 1: That sounds really interesting! \\
    Speaker 2: I had a lot of fun writing it. \\
    Speaker 1: Do you plan on doing anything with the story, like submitting it for \\ publication or sharing it with friends? \\
    Speaker 2: I'm considering submitting it to some literary magazines, but I haven't \\
    decided yet. \\
    Speaker 1: That sounds like a good plan. \\
    Speaker 2: I might also share it with some friends to get their feedback. \\
    Speaker 1: I'm sure it will be well-received. \\
    \midrule
    \end{tabular}
    \caption{Session 4 of an sample conversation we collect.}
    \label{tab:sample_data_4}
\end{table*}

\section{Instruction and interface for data collection}
\label{apx_c}

\begin{figure*}[t!]
    \centering
    \begin{subfigure}[b]{0.8\textwidth}
        \includegraphics[width=0.9\textwidth]{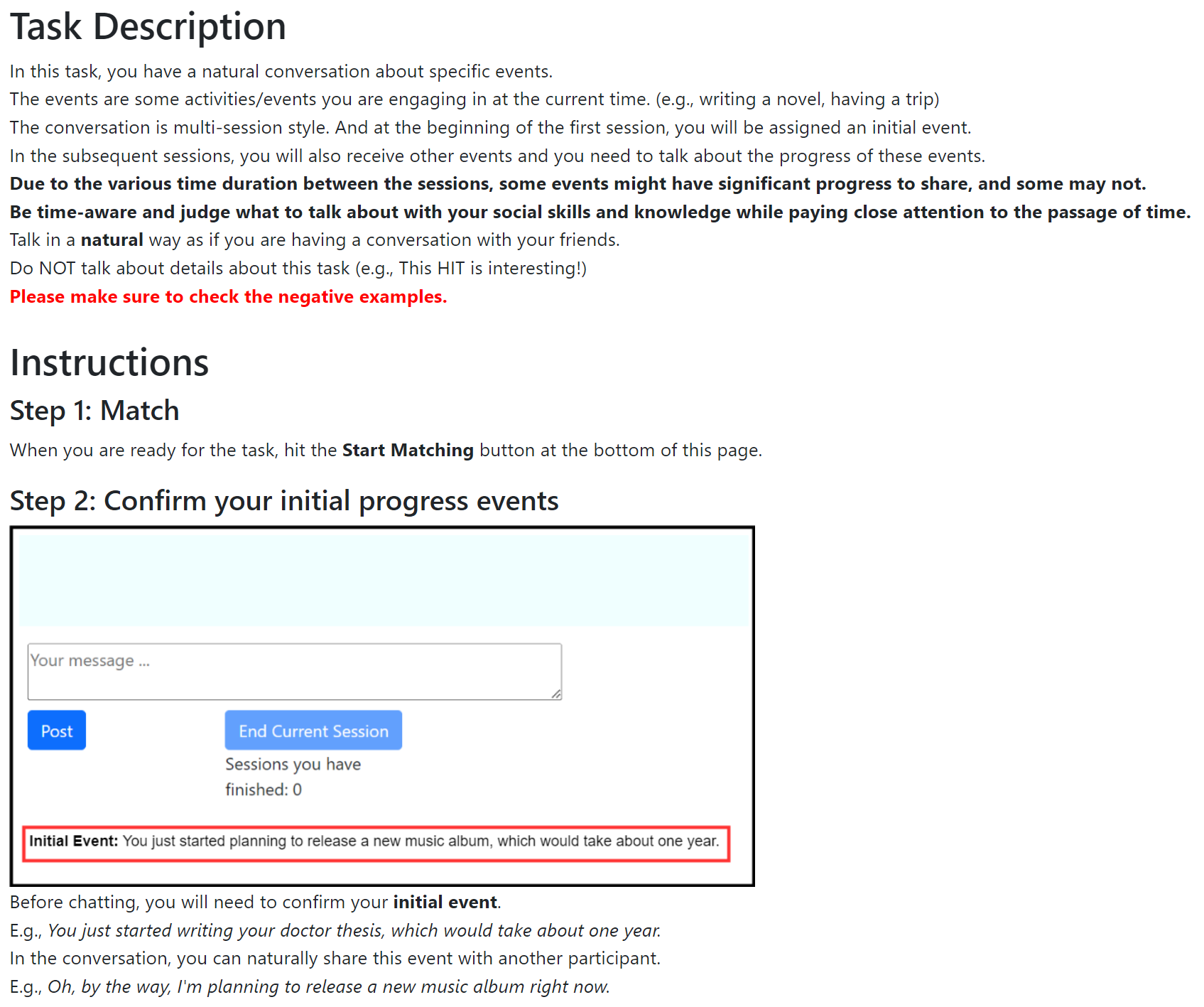}
    \end{subfigure}
    \begin{subfigure}[b]{0.8\textwidth}
        \includegraphics[width=0.9\textwidth]{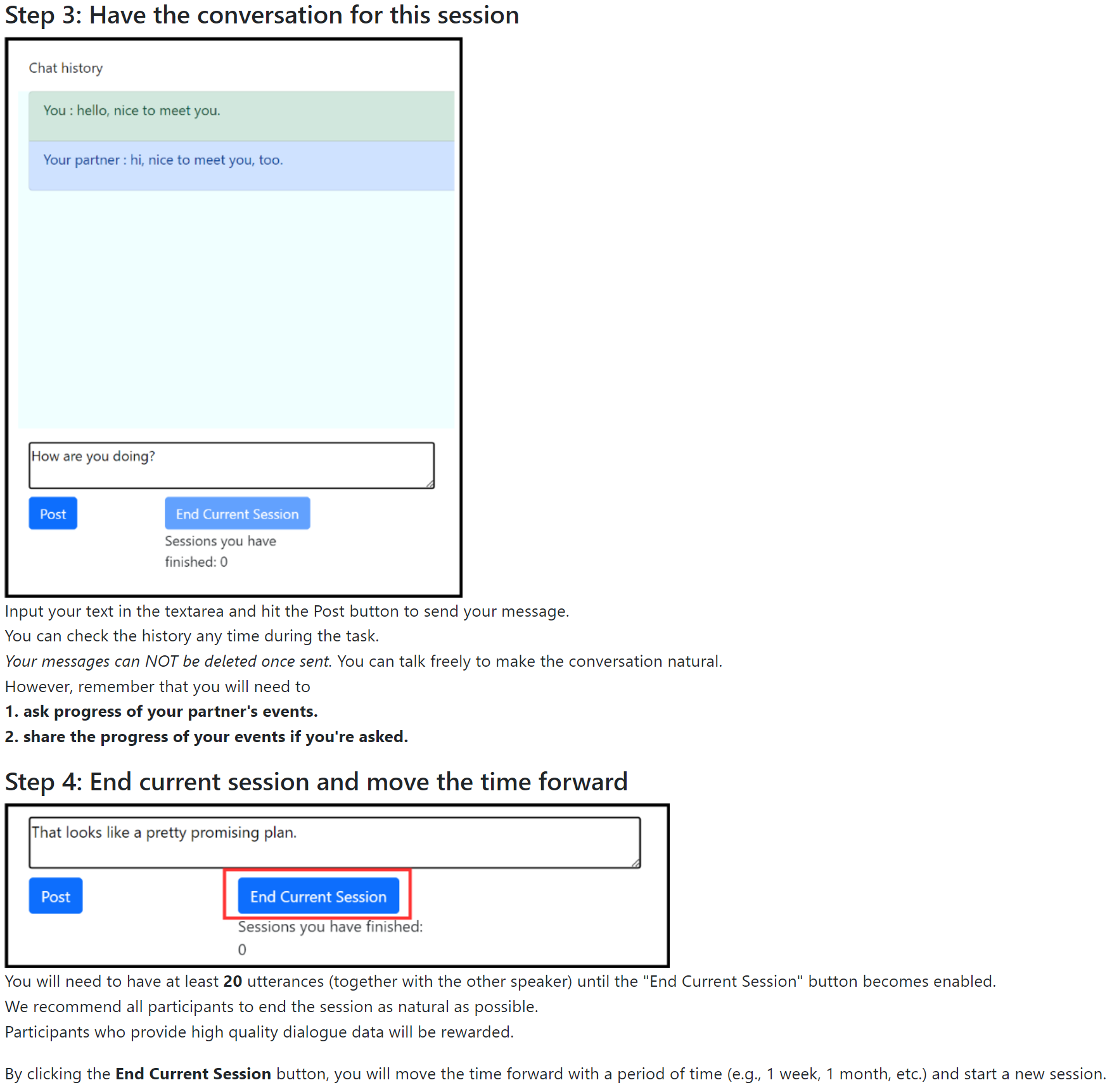}
    \end{subfigure}
    \caption{The instruction we use for data collection.}
    \label{fig:instruction_1}
\end{figure*}

\begin{figure*}
    \centering
    \begin{subfigure}[b]{0.8\textwidth}
        \includegraphics[width=0.9\textwidth]{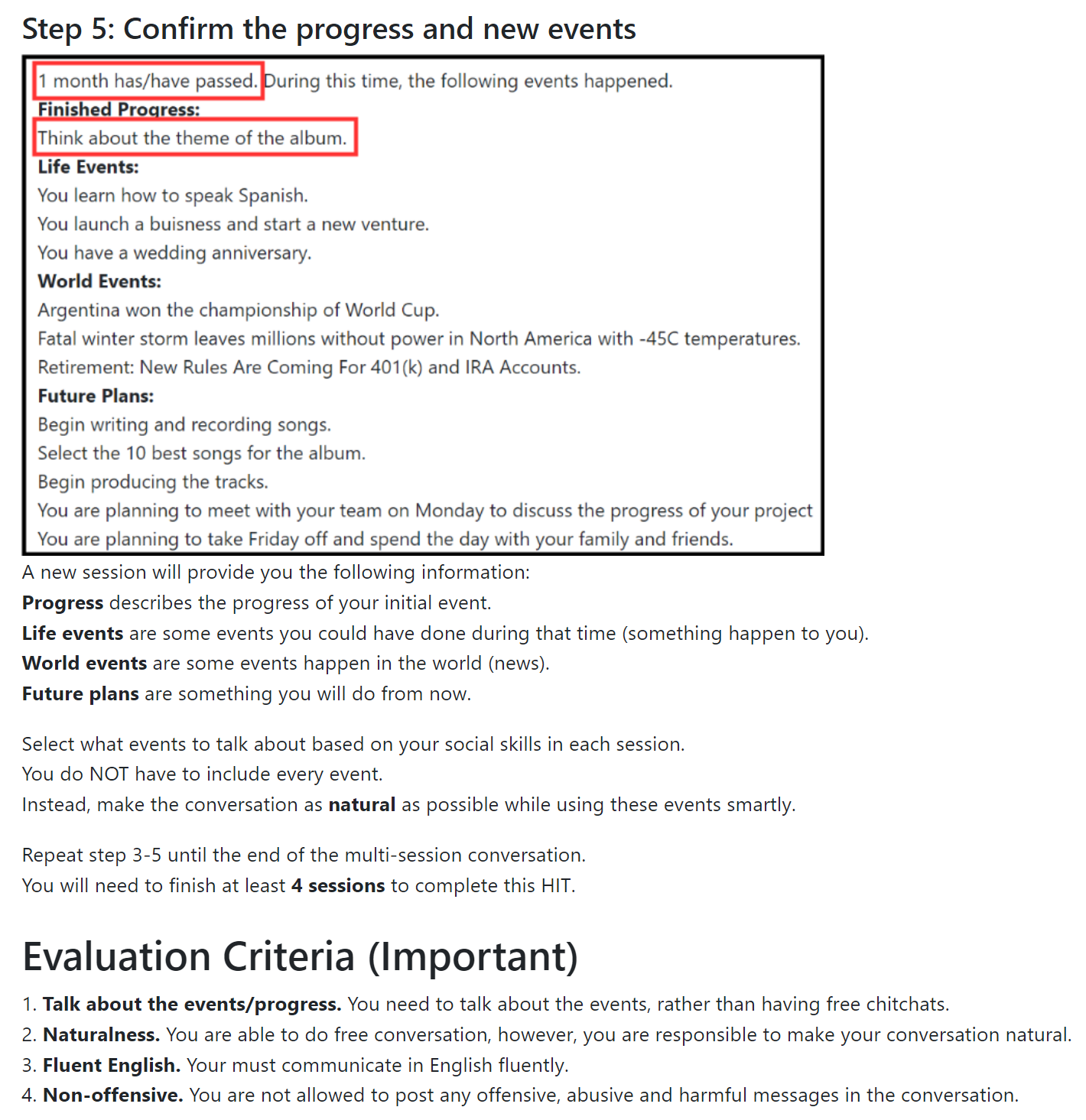}
    \end{subfigure}
    \begin{subfigure}[b]{0.8\textwidth}
        \includegraphics[width=0.9\textwidth]{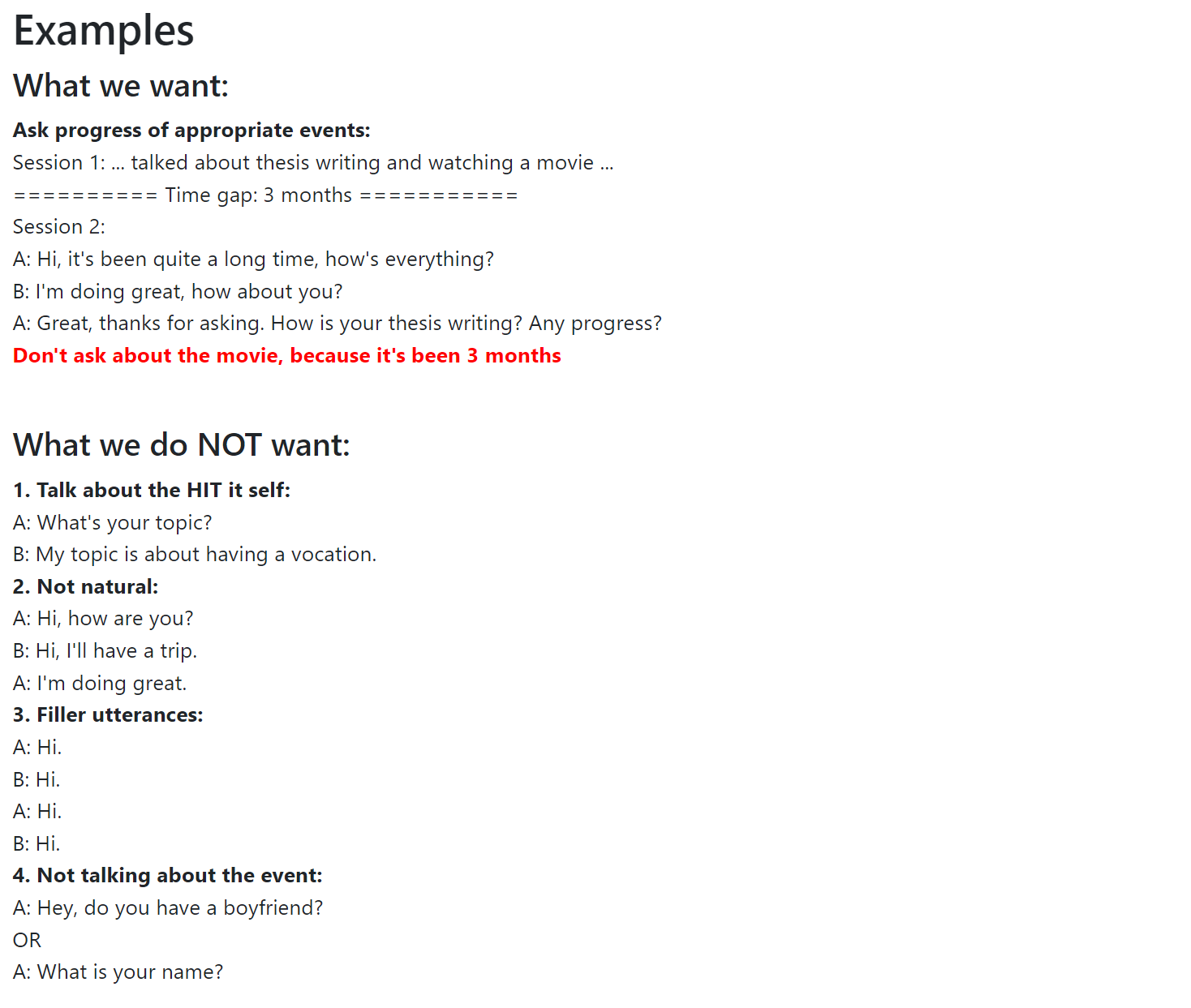}
    \end{subfigure}
    \caption{The instruction we use for data collection.}
    \label{fig:instruction_2}
\end{figure*}

Figure~\ref{fig:instruction_1} and \ref{fig:instruction_2} show the instructions we use for data collection. We explicitly ask speakers to follow the events and timelines in the instruction. We also raise the point that any offensive, abusive content will not be allowed and the speakers are able to report and stop and task any time. We also provide positive and negative examples to demonstrate how the data should look like for the speakers.

\begin{figure*}[t!]
    \centering
    \includegraphics[width=0.8\textwidth]{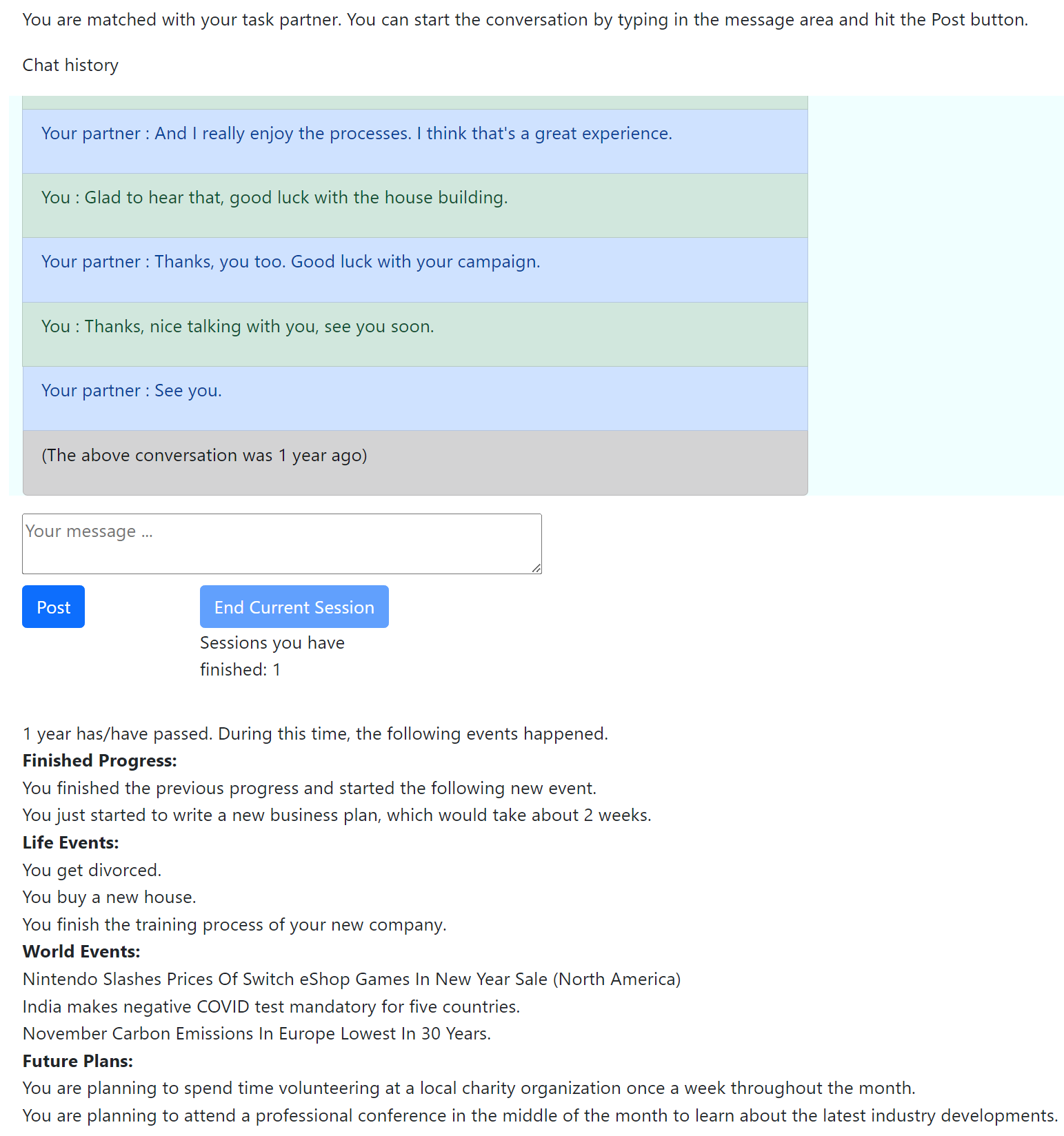}
    \caption{The interface for speakers to have conversations.}
    \label{fig:interface}
\end{figure*}

To generate the conversations, we use a matching system to first randomly match two participants. After matching, the participants are redirected to the chat room as shown in Figure~\ref{fig:interface}.
At the beginning of the conversation, each participant will be shown their initial events for the participants to talk about. When the session reaches the maximum length, either participant could move the time forward for a random time gap by clicking the ``End current session'' button. And for the new session, we show different types of information to both participants. Finished progress indicates the life events they were engaging in the previous session. The progress is represented as the schedule steps based on the time gap. We also show some random life events and world events defined in the timeline. Finally, we also provide some future plans for the participants. The future plans are either some future schedule steps for unfinished progress or new events in the timeline if the previous events are finished.

\section{Prompts for modeling time}
\label{apx_d}
\begin{figure}
    \centering
    \includegraphics[width=0.48\textwidth]{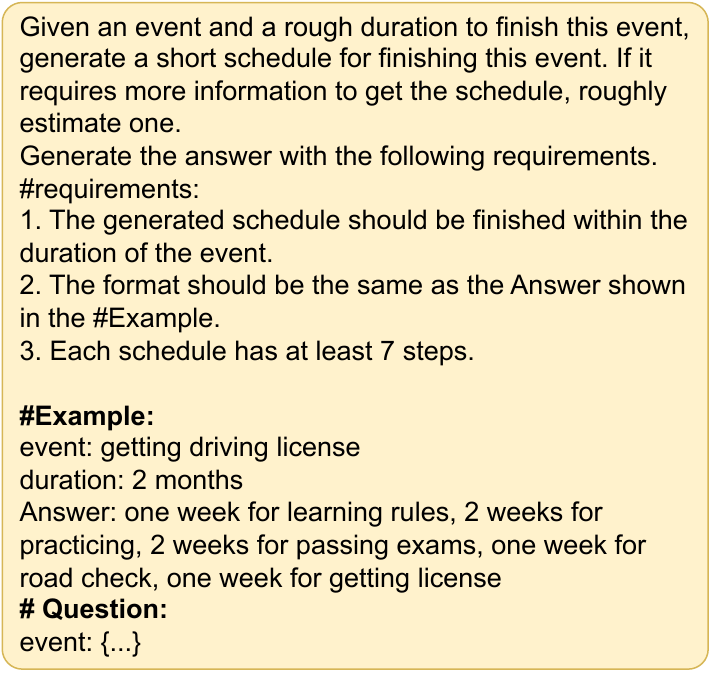}
    \caption{The prompt we use to generate the steps towards finishing a life event.}
    \label{fig:craft_events}
\end{figure}
\begin{figure}[t]
    \centering
    \includegraphics[width=0.48\textwidth]{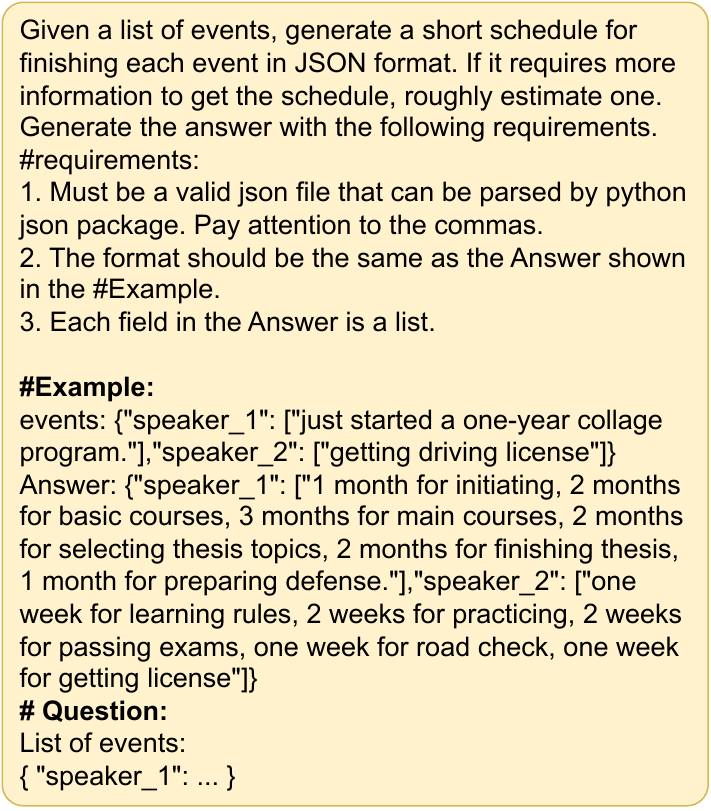}
    \caption{The prompt we use to get schedules for the pre-defined events during training and extracted events during inference.}
    \label{fig:get_schedule}
\end{figure}

\subsection{Prompts we used in our experiment settings}

Figure~\ref{fig:craft_events} is the prompt we use to generate the steps (schedule) towards finishing a life event in our training data.

Figure~\ref{fig:get_schedule} is the prompt we use to generate short schedules for given events. These prompts are used as they show best performance in our tests.

\subsection{Prompts we tried with different styles}
We try different types of prompts following previous works researching on the factors that could affect the performance of prompting results~\cite{schick-schutze-2021-exploiting,lu-etal-2022-fantastically,mishra-etal-2022-reframing}. Only few-shot prompting method is explored because we require the generated events to be in certain format (Table~\ref{tab:completion}$\sim$\ref{tab:qa_2}).

For the question-answering style, we only fix the first two questions and dynamically select the questions based on the answer to previous question. For instance, if the answer to Question 1 is ``Yes'', we will choose ``What are the events that speaker A is engaging?'' as Question 3. When asking a question, all the question and answers to previous questions are included as part of the prompt.

\begin{table*}[t]
    \centering
    \begin{tabular}{l}
        \toprule
        \textit{Instruction description} \\
        \midrule
        In the following conversation, the speakers are engaging in some events that take a certain amount  \\ 
        of time. Extract such events and estimate the expected time to finish these events. \\
        \midrule
        \textit{Instances $*N$} \\
        \midrule
        \textbf{Conversation:} \\
        A: Hi how are you? \\
        B: Yes I am fine and how are you doing today? \\
        A: Doing good. What is the plan for tonight? \\
        B: Not yet planed for something. I just started with preparing and executing a social media  \\
        marketing campaign for my company. \\
        A: Oh are you busy in that? \\
        \textbf{Events:} \\
        B: executing a social media marketing (about 3 months) \\
        \midrule
        \textit{Question} \\
        \midrule
        A: Hi, how are you doing? \\
        B: I'm doing great, how about you? \\
        A: I'm also doing good. I'm just busy with my paper writing as the deadline is approaching. \\
        \textbf{Events:} \\
        <extracted events> \\
    \bottomrule
    \end{tabular}
    \caption{An example of few-shot prompting with information explaining }
    \label{tab:completion}
\end{table*}

\begin{table*}[t]
    \centering
    \begin{tabular}{l}
        \toprule
        \textit{Instances $*N$} \\
        \midrule
        \textbf{Conversation:} \\
        A: Hi how are you? \\
        B: Yes I am fine and how are you doing today? \\
        A: Doing good. What is the plan for tonight? \\
        B: Not yet planed for something. I just started with preparing and executing a social media  \\
        marketing campaign for my company. \\
        A: Oh are you busy in that? \\
        \textbf{Events:} \\
        In the above conversation, speakers talked about the events they are engaging. A is engaging in \\
        \textbf{something is not mentioned}. B is engaging in \textbf{executing a social media marketing, } \\
        \textbf{which takes about 3 months.} \\
        \midrule
        \textit{Question} \\
        \midrule
        A: Hi, how are you doing? \\
        B: I'm doing great, how about you? \\
        A: I'm also doing good. I'm just busy with my paper writing as the deadline is approaching. \\
        \textbf{Events:} \\
        In the above conversation, speakers talked about the events they are engaging. \_\_\_\_ is engaging in \\
        \_\_\_\_. \_\_\_\_ is engaging in \_\_\_\_. \\
        \bottomrule
    \end{tabular}
    \caption{An example of few-shot prompting for slot filling, where the LLM is required to fill in the blanks according to given instances.}
    \label{tab:qa_1}
\end{table*}

\begin{table*}[t]
    \centering
    \begin{tabular}{l}
        \toprule
        \textit{Instances $*N$} \\
        \midrule
        \textbf{Conversation:} \\
        \midrule
        A: Hi how are you? \\
        B: Yes I am fine and how are you doing today? \\
        A: Doing good. What is the plan for tonight? \\
        B: Not yet planed for something. I just started with preparing and executing a social media  \\
        marketing campaign for my company. \\
        A: Oh are you busy in that? \\
        \midrule
        \textit{Questions} \\
        \midrule
        \textbf{Question 1:} \\
        Did speaker A mention any events that speaker A is engaging? Answer with Yes or No\\
        Answer: \_\_\_\_ (No) \\
        \textbf{Question 2:} \\
        Did speaker B mention any events that speaker B is engaging?
        Answer with Yes or No \\
        Answer: \_\_\_\_ (Yes) \\
        \textbf{Question 3:} \\
        What are the events that speaker B is engaging?
        Answer the the content of the event and an \\
        estimated time to finish that event. \\
        Answer: Speaker B is engaging in executing a social media marketing, which takes about 3 months \\
        to finish. \\
        \textbf{Question 4:} \\
        Give a rough schedule of the events that B is engaging within the estimated time. \\
        Answer: \\
        Week 1-4: Research the target audience and their social media habits. Identify the platforms \\ 
        that the target audience uses most.\\
        Week 5: Create high-quality, engaging content that aligns with the campaign goals and appeals to \\
        the target audience.\\
        ... \\
        \bottomrule
    \end{tabular}
    \caption{An example of question answering style prompt to extract the events and their duration.}
    \label{tab:qa_2}
\end{table*}
\clearpage
\section{Prompt for collecting conversations from ChatGPT}
\label{apx_e}
\begin{figure}[t]
    \centering
    \includegraphics[width=0.48\textwidth]{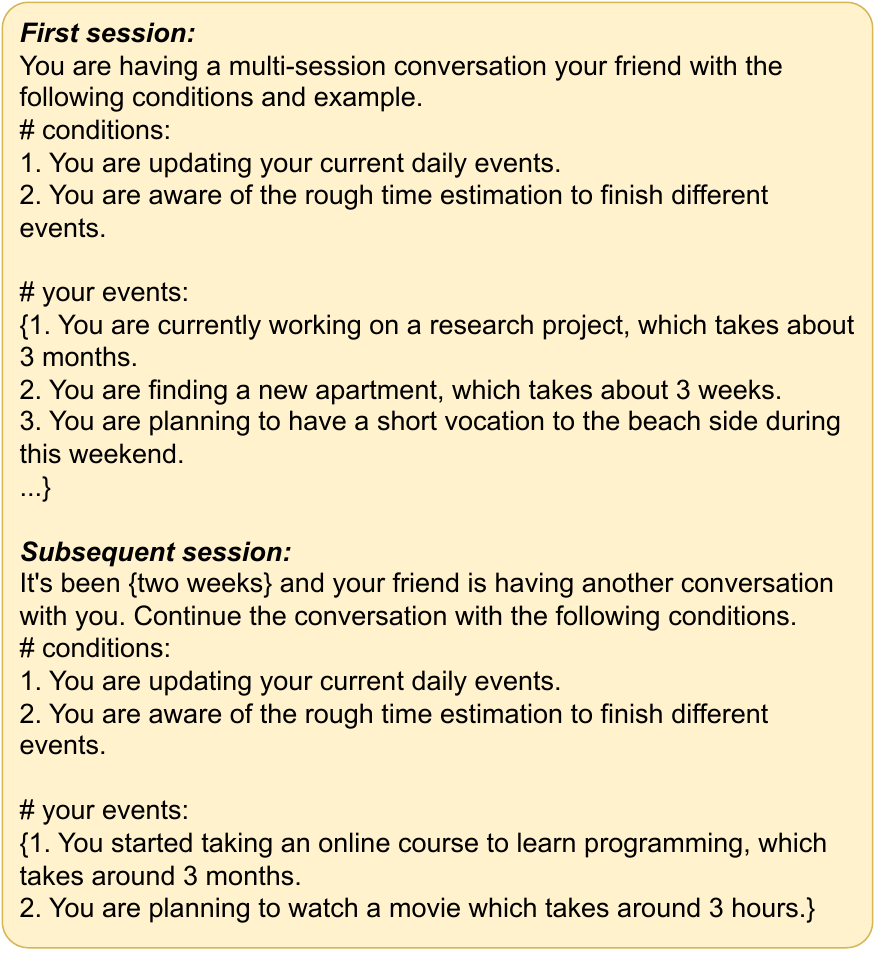}
    \caption{Prompts for generating conversations from ChatGPT (with gap). Content in \{\} is replaced with same events and time gaps as in the collection process of other RAG models.}
    \label{fig:chatgpt_prompt}
\end{figure}

Figure~\ref{fig:chatgpt_prompt} is the prompt we use for collecting conversations from ChatGPT. We use two prompts to generate the first session and subsequent sessions. For the first session, we prompt ChatGPT to have a multi-session conversation with the same event settings as in other TA-RAG models. This prompt contains the conditions and events information. After generating the first session, we use a different prompt for generating the subsequent sessions. In the time-aware case of this prompt, we explicitly tell ChatGPT that there is a time gap between this session and the previous session and previous dialogue history is included as part of the prompt. First several utterances will be removed when the conversation is longer than the maximum input of ChatGPT. When generating time-aware ChatGPT conversations, we add another section of progress labels or schedules into the prompt we use for generating subsequent sessions. 


\section{Human Evaluation Questionnaire}
\label{apx_f}

We ask the annotators in total 11 questions (Table~\ref{tab:questionnaire}) and ask them to choose one system over another one. For question Q3, Q6, Q9 and Q11 we ask annotators to provide justifications to explain the reason of their choices. The justifications are used for both filtering out invalid answers and analyzing users feedback. We consider the following two types of annotations as invalid.

First, we remove those annotations that are with an extremely short working time. Given the length of the dialogues and questions, we consider it impossible to finish the annotation within 200 seconds. Therefore, any annotations that are with a work duration less than 200 seconds are directly filtered out. Second, we remove those annotations that are with unreasonable justifications. For instance annotators may give [``good'', ``good'', ``good'', ``good''] as justifications, which are obvious signs of the worker not engaging in the task. We perform pilot tests to select good annotators for our tasks. The annotators are required to have finished more than 2000 tasks, with an approval rate over 97\% and be fluent in English.

\begin{table*}[t]
    \centering
    \begin{tabular}{l}
    \toprule
    \textbf{Questions} \\
    \midrule
    \textit{Naturalness} \\
    \midrule
    Q1. Which dialogue do you think is more natural like two friends updating their daily life?   \\
    Q2. Which dialogue do you think is more like a dialogue between normal friends? \\
    Q3. In which dialogue do you think the speaker B talks more naturally? \\
    \midrule
    \textit{Informativeness} \\
    \midrule
    Q4. Which dialogue do you think provide more information about the speakers' daily events?    \\
    Q5. In which dialogue do you think the speaker B asks annoying questions more \\ 
    frequently about events that have no significant progress? \\
    Q6. In which dialogue do you think the speaker B cares more about speaker A's daily events?\\
    \midrule
    \textit{Relevance} \\
    \midrule
    Q7. Does the speaker asks/talks about relevant events? \\
    Q8. Which dialogue do you think sticks to the topics of updating events? \\
    Q9. In which dialogue do you think the speaker chooses natural, relevant events to talk about? \\
    \midrule
    \textit{Time-awareness} \\
    \midrule
    Q10. In which dialogue do you think speaker B can identify a time gap? \\
    Q11. In which dialogue do you think speaker B is aware of the progress of relevant events?\\
    \bottomrule
    \end{tabular}
    \caption{The questions we ask annotators for the human evaluation.}
    \label{tab:questionnaire}
\end{table*}

\section{Sample Conversations}
\label{sec:apx_g}

Table~\ref{tab:sample_dialogue_1}$\sim$\ref{tab:sample_dialogue_3} show an example of the conversations generated by TA-RAG (both) (Speaker B).
\newpage

\begin{table*}[t]
    \centering
    \begin{tabular}{l}
    \midrule
    Events: \\
    A: You watch a movie. \\
    You are going to visit your grandparents. \\
    B: You plan to go to a music festival. \\
    \midrule
    A: I am doing well. I love the outdoors as well. What beach did you go to? \\
    B: do you have any plans for the upcoming weekend? I'm planning to go to a festival. \\
    A: That sounds like fun. I am going to a concert this weekend. What kind of music do you \\
    listen to? \\
    B: What about you? What do you do for fun? I like to dance and go to concerts. \\
    A: I like a lot of different kinds of music. My favorite band is Led Zeppelin. \\
    B: what is your favorite thing to do in the evening? I love to go dancing with my friends. \\
    A: I like going out to eat with my family. We like to eat out a lot. Do you have a favorite\\
    food? \\
    B: That's great. I also like to go out with friends. I like dancing and dancing. \\
    A: Dancing is a great way to get out and have fun. What type of dancing do you enjoy? \\
    B: I love dancing, but I don't dance much. I'm more of an outdoor person. \\
    A: Outdoor dancing is great too. It is so much fun to be out in nature. \\
    B: I love that movie. It's one of my all time favorites. Have you seen it? \\
    A: No, I haven't seen it. I have heard it is good though. What is it about? \\
    B: I love watching movies and listening to music. What's your favorite genre of music to \\
    dance to? \\
    A: I love pop music. I listen to it all the time while I work out at the gym. \\
    B: That's nice. What about your family? Are you planning to do anything special for the \\
    holidays? \\
    A: My family is great. We are going to visit my grandparents. They live in Florida. \\
    B: Do you like to watch movies or tv shows? I am a big fan of the genre. \\
    A: I like movies and tv shows too. What are some of your favorite shows to watch on tv? \\
    B: I'm going to watch a movie with my brothers and sisters. It will be fun. \\
    A: What movie are you going to see with your siblings? I hope it's a good movie. \\
    \bottomrule
    \end{tabular}
    \caption{Session 1 of an conversation generated by TA-RAG (both).}
    \label{tab:sample_dialogue_1}
\end{table*}

\begin{table*}[t]
    \centering
    \begin{tabular}{l}
    \midrule
    Gap: 1 week \\
    Events: \\
    A: You watch tv dramas. \\
    B: You plan to go hiking. \\
    \midrule
    B: what about you what are your plans for this weekend? What are you up to? I plan to go\\
    hiking. \\
    A: Hiking sounds like a good idea. I might go for a hike too. I've never been hiking before. \\
    B: how were your grandparents? are they doing great? \\
    A: Thanks for asking, they are doing great. And I had a great time with them. We watched \\
    Game of Thrones. \\
    B: that's great, I'm glad to hear that. How about you, what's going on? \\
    A: I just finished watching the last season of House of Cards. It was a great show. \\
    B: yes, I like it too, it is a romantic comedy and romance comedy. \\
    A: It really is. I can't wait for the new season to come out in 2019. \\
    B: do you know any good romance movies? I really like romance and thrillers movies. \\
    A: I do! I love The Notebook. I think it's my favorite movie of all time. \\
    B: I love romantic movies, especially romantic ones. I just watched a movie about romance. \\
    A: That sounds interesting. I'll have to check it out. What did you think of the movie? \\
    B: I like romantic drama and thriller movies, do you also like those genres of movies? \\
    A: I do like those types of movies. I especially like the thrill of a good thriller. \\
    B: yes, i love romance, action, adventure, thriller, mystery, suspense. \\
    A: Have you ever seen the movie "The Sixth Sense" by David Suchet? \\
    \bottomrule
    \end{tabular}
    \caption{Session 2 of an conversation generated by TA-RAG (both).}
    \label{tab:sample_dialogue_2}
\end{table*}

\begin{table*}[t]
    \centering
    \begin{tabular}{l}
    \midrule
    Gap: 2 months \\
    Events: \\
    A: You prepare for the coming tests. \\
    B: You just started to learn to play piano, which takes about 5 months. \\
    \midrule
    B: I'm doing great, thanks for asking. How is your day going? What have you been up to\\
    lately? \\
    A: My day is going well, thank you! I've been studying for a test I have tomorrow. \\
    B: what are you planing to do for the weekend? I am planning to see a movie.  \\
    A: I plan on going to the movies as well! I'm looking forward to seeing the new Star Wars movie! \\
    B: do you want to go with me? I can take you to the theater to see the movie. \\
    A: That would be great! I'll bring my boyfriend as well, he's the only one I've ever gone with. \\
    B: I like watching movies, tv shows, documentaries, documentarys, tv series and documentaries. \\
    A: I love documentaries! I watch them every night before I go to bed. Whats your favorite? \\
    B: do you have any other hobbies or hobbies? I just started learning how to play the piano. \\
    A: That's so cool! I play the drums. I started when I was a teenager. \\
    B: I'm planning to take a trip to the beach in the next few months. I hope you can join me. \\
    A: I would love to! I have never been to a beach before. Where will you be going? \\
    B: It is a place somewhere near to the south east coast.\\
    A: That sounds like a lot of fun. I hope I could go together with you. \\
    \bottomrule
    \end{tabular}
    \caption{Session 3 of an conversation generated by TA-RAG (both).}
    \label{tab:sample_dialogue_3}
\end{table*}
\clearpage
\section{TA-RAG model's performance over different session gaps}
\label{apx_h}

\begin{figure}
    \centering
    \includegraphics[width=0.48\textwidth]{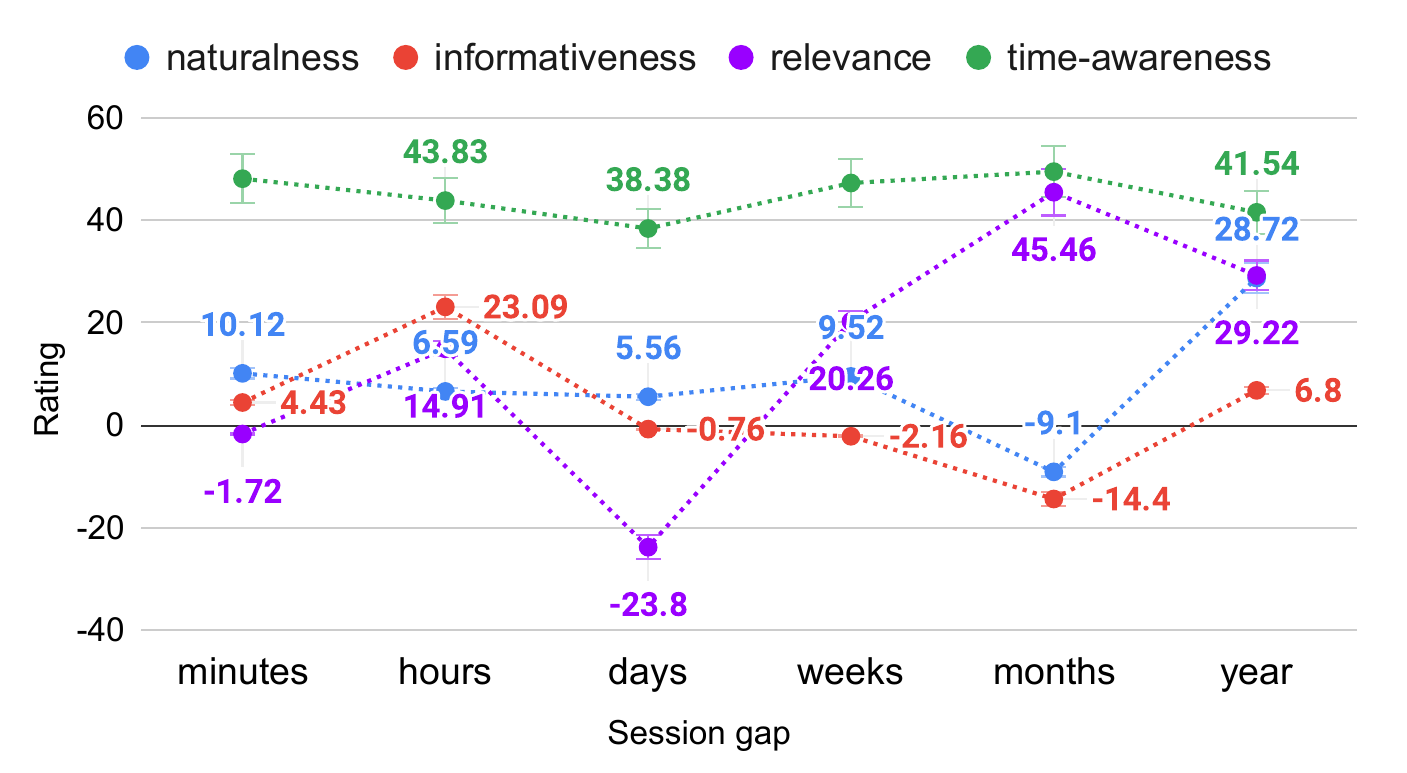}
    \caption{Performance of TA-RAG (progress) over different session gaps.}
    \label{fig:session_gap_time}
\end{figure}

\begin{figure}
    \centering
    \includegraphics[width=0.48\textwidth]{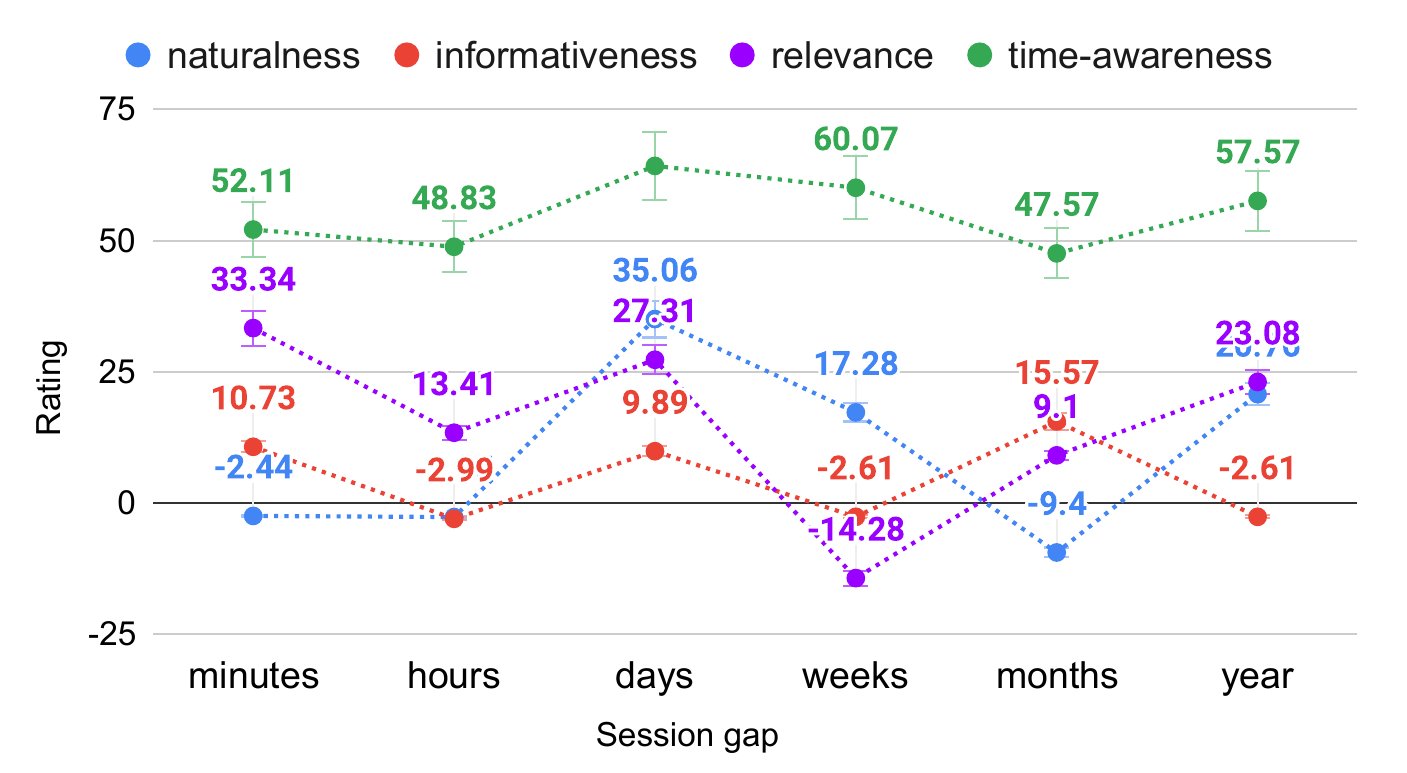}
    \caption{Performance of TA-RAG (schedule) over different session gaps.}
    \label{fig:session_gap_schedule}
\end{figure}

Table~\ref{tab:session_gap} shows the numbers of different session gaps in evaluation experiments.
Figure~\ref{fig:session_gap_time} and~\ref{fig:session_gap_schedule} are the performance of TA-RAG models with different time-aware information over different session gaps. Although we are able to observe more stable naturalness for TA-RAG (both) model, that does not hold in TA-RAG (schedule). Informativeness is also more stable in TA-RAG (progress), however shows larger deviation for TA-RAG (both) and TA-RAG (schedule).
\begin{table}[h!]
    \centering
    \begin{tabular}{lc}
    \toprule
    Session gap & \#sessions \\
    \midrule
         minutes &  82\\
         hours &  90 \\
         days &  65 \\
         weeks & 78 \\
         months & 62 \\ 
         year & 73 \\
    \bottomrule
    \end{tabular}
    \caption{Number of different session gaps in our experiments.}
    \label{tab:session_gap}
\end{table}

\end{document}